\documentclass[conference,final]{IEEEtran}

\usepackage[colorlinks=true,linkcolor=black,anchorcolor=black,citecolor=black,filecolor=black,menucolor=black,runcolor=black,urlcolor=black]{hyperref}
\usepackage{subcaption}
\usepackage{gensymb}
\usepackage{booktabs} 
\usepackage[colorinlistoftodos]{todonotes}
\usepackage[backend=biber]{biblatex}
\begin{document}

\title{Heritability in Morphological Robot Evolution}

\author{\IEEEauthorblockN{Matteo De Carlo}
\IEEEauthorblockA{Vrije Universiteit Amsterdam\\
Amsterdam, Netherlands\\
m.decarlo@vu.nl}
\and
\IEEEauthorblockN{Eliseo Ferrante}
\IEEEauthorblockA{
\textit{Vrije Universiteit Amsterdam}\\
Amsterdam, Netherlands \\
\textit{Technology Innovation Institute} \\
\hspace{1.0cm}Abu Dhabi, United Arab Emirates\hspace{1.0cm} \\
e.ferrante@vu.nl}
\and
\IEEEauthorblockN{Daan Zeeuwe}
\IEEEauthorblockA{Vrije Universiteit Amsterdam\\
Amsterdam, Netherlands\\
d.zeeuwe@vu.nl}
\and
\IEEEauthorblockN{Jacintha Ellers}
\IEEEauthorblockA{Vrije Universiteit Amsterdam\\
Amsterdam, Netherlands\\
j.ellers@vu.nl}
\and
\IEEEauthorblockN{Gerben Meynen}
\IEEEauthorblockA{Vrije Universiteit Amsterdam\\
Amsterdam, Netherlands\\
g.meynen@vu.nl}
\and
\IEEEauthorblockN{A.E.Eiben}
\IEEEauthorblockA{Vrije Universiteit Amsterdam\\
Amsterdam, Netherlands\\
a.e.eiben@vu.nl}}

\maketitle

\begin{abstract}
In the field of evolutionary robotics, choosing the correct encoding is very complicated, especially when robots evolve both behaviours and morphologies at the same time. 
With the objective of improving our understanding of the mapping process from encodings to functional robots, we introduce the biological notion of heritability, which captures the amount of phenotypic variation caused by genotypic variation. 
In our analysis we measure the heritability on the first generation of robots evolved from two different encodings, a direct encoding and an indirect encoding.
In addition we investigate the interplay between heritability and phenotypic diversity through the course of an entire evolutionary process. 
In particular, we investigate how direct and indirect genotypes can exhibit preferences for exploration or exploitation throughout the course of evolution.
We observe how an exploration or exploitation tradeoff can be more easily understood by examining patterns in heritability and phenotypic diversity.
In conclusion, we show how heritability can be a useful tool to better understand the relationship between genotypes and phenotypes, especially helpful when designing more complicated systems where complex individuals and environments can adapt and influence each other.
\end{abstract}

\IEEEpeerreviewmaketitle

\section{Introduction}

Evolutionary robotics (ER) employs several elements of biological evolution to obtain creative and novel solutions to practical problems.
A key concept in evolutionary biology that could help in designing evolutionary robotic systems and that so far has remained largely unexplored in evolutionary robotics is the notion of heritability~\cite{griffiths_quantifying_2000, wray2008estimating}.
Heritability is one of the factors determining the ability of traits to evolve.
A common use of heritability is in animal and plant breeding, where the response to selection can be predicted as the product of heritability and the selection differential~\cite{lynch1998genetics}.

Heritability denotes the proportion of additive genetic variance relative to the total phenotypic variance, and can vary between zero and unity. 
Hence it indicates the degree to which the trait is responsive to selection. 
Low heritabilities are often associated with a more diffuse genotype-phenotype map, possibly resulting from a strong influence of the environment on the phenotype or interaction among genes in their expression (epistasis).
Because low heritability compromises the evolutionary response, the notion of heritability can be a useful addition to evolutionary robotics as an a priori evaluation of the evolutionary potential of a system.
 
One important design choice that has a major impact but is often overlooked when designing an ER system is the choice of encoding.
There are several encodings that have been used in the literature available to the designer, and many more if one includes all variations.
Despite so, our understanding on how the choice on a particular encoding can influence the evolutionary process is still very superficial.
The main goal of this paper is to investigate the applicability of the notion of heritability in an evolutionary robotics system to better understand the relationship between different encodings and the generated phenotypes.

Heritability can be defined for any given phenotypic trait of the robots, either related to the robot's morphology, (i.e. size), or the robot's behavior (i.e. speed). Then for any pair of parent robots, we can determine the average value of the given trait and compare it with the value of this trait in the child robot. It is important to note that while the traits we consider are phenotypic, the mechanisms that transfer them to the offspring depend on the genotypes, specifically, the genetic encoding that specifies that trait and the recombination operator that shuffles and combines the parental genotypes into a new one that represents the child. In principle, this means that based on observable phenotypic properties we can get information regarding genotypic processes ``under the hood''.

The main contribution is the adoption of the concept of heritability to Evolutionary Robotics and the demonstration of its utility. 
There are three important aspects we investigate:
\begin{itemize}
    \item Whether heritability can be used as a predictor of the evolutionary response of a system, specifically whether it is related to the rate of evolutionary change.
    \item Whether heritability changes over the course of evolution and, if so, if this effect can be related to any other measurable aspect of evolution. 
    \item Whether heritability can be used to evaluate and compare different genetic representations.
\end{itemize}


%


\section{System description}

\subsection{The Robots}
\label{sec:robotsystem}


\begin{figure}[t]
    \centering
    \begin{subfigure}[b]{0.37\linewidth}
        \includegraphics[clip,width=\textwidth]{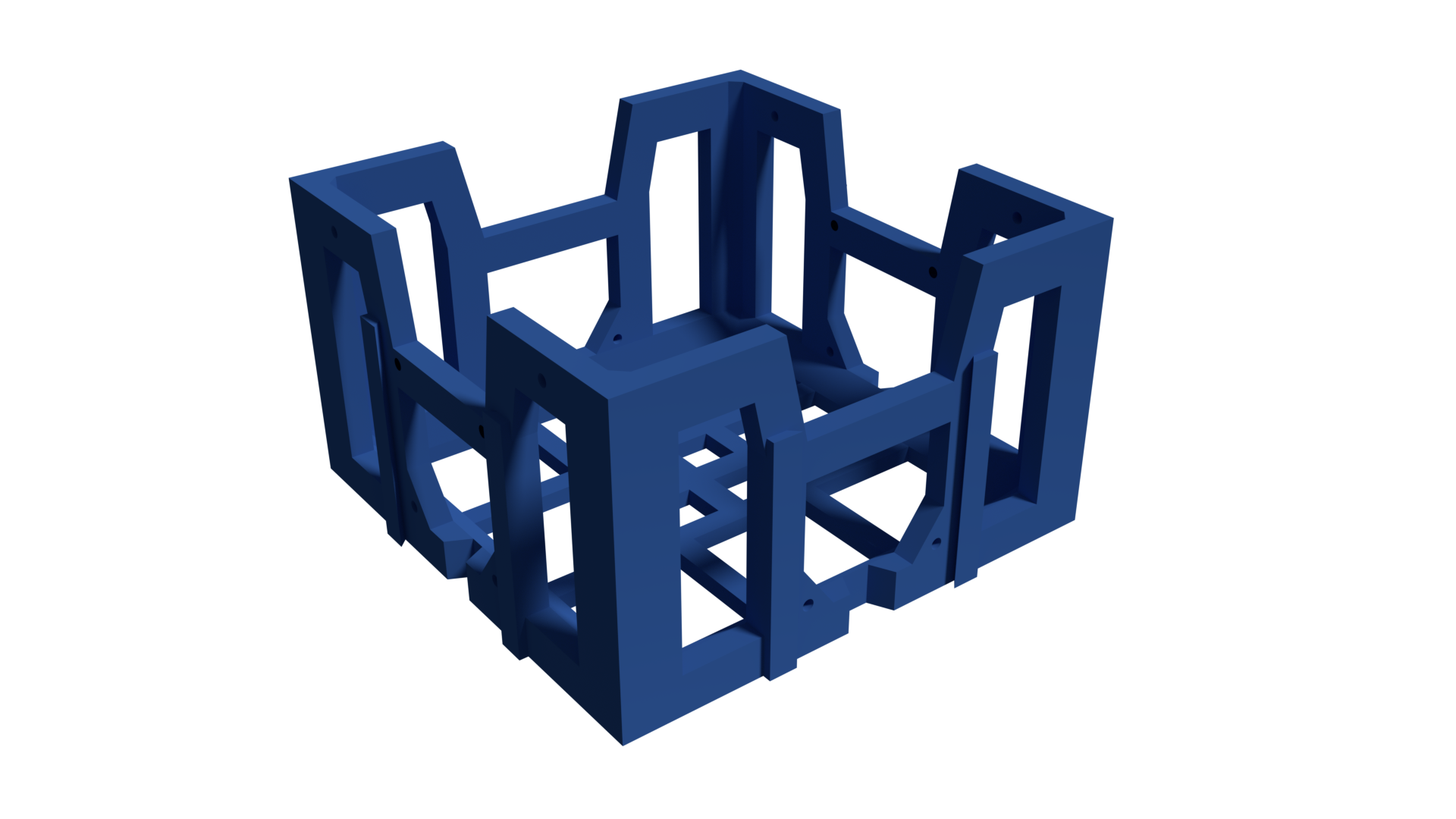}
        \caption{Core module}
        \label{fig:robot_modules:core}
    \end{subfigure}
    \begin{subfigure}[b]{0.3\linewidth}
        \includegraphics[trim={10cm 0 -4cm 0},clip,width=\textwidth]{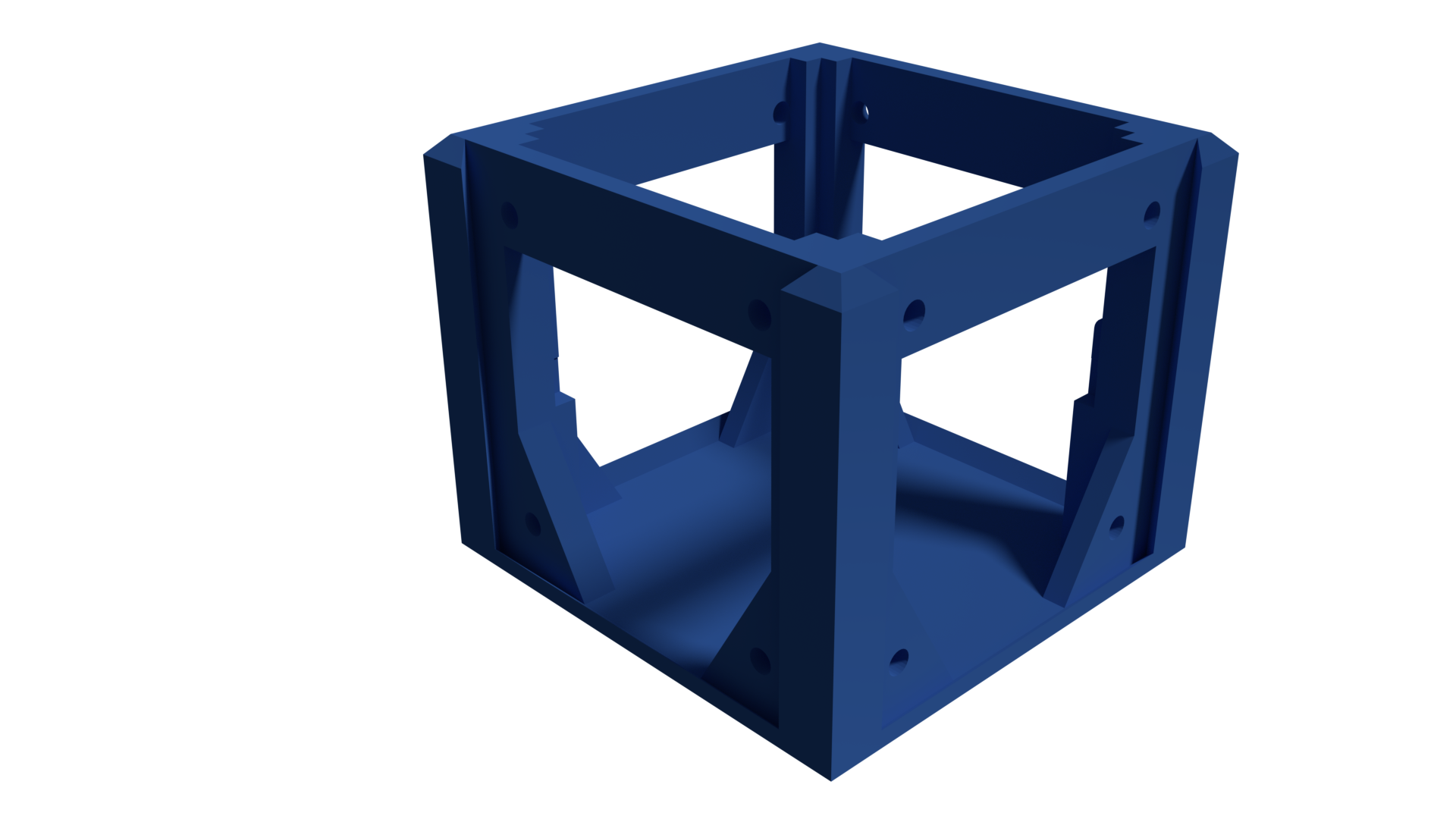}
        \caption{Brick module}
        \label{fig:robot_modules:brick}
    \end{subfigure}
    \begin{subfigure}[b]{0.3\linewidth}
        \includegraphics[trim={8cm 4cm 9cm 4cm},clip,width=\textwidth]{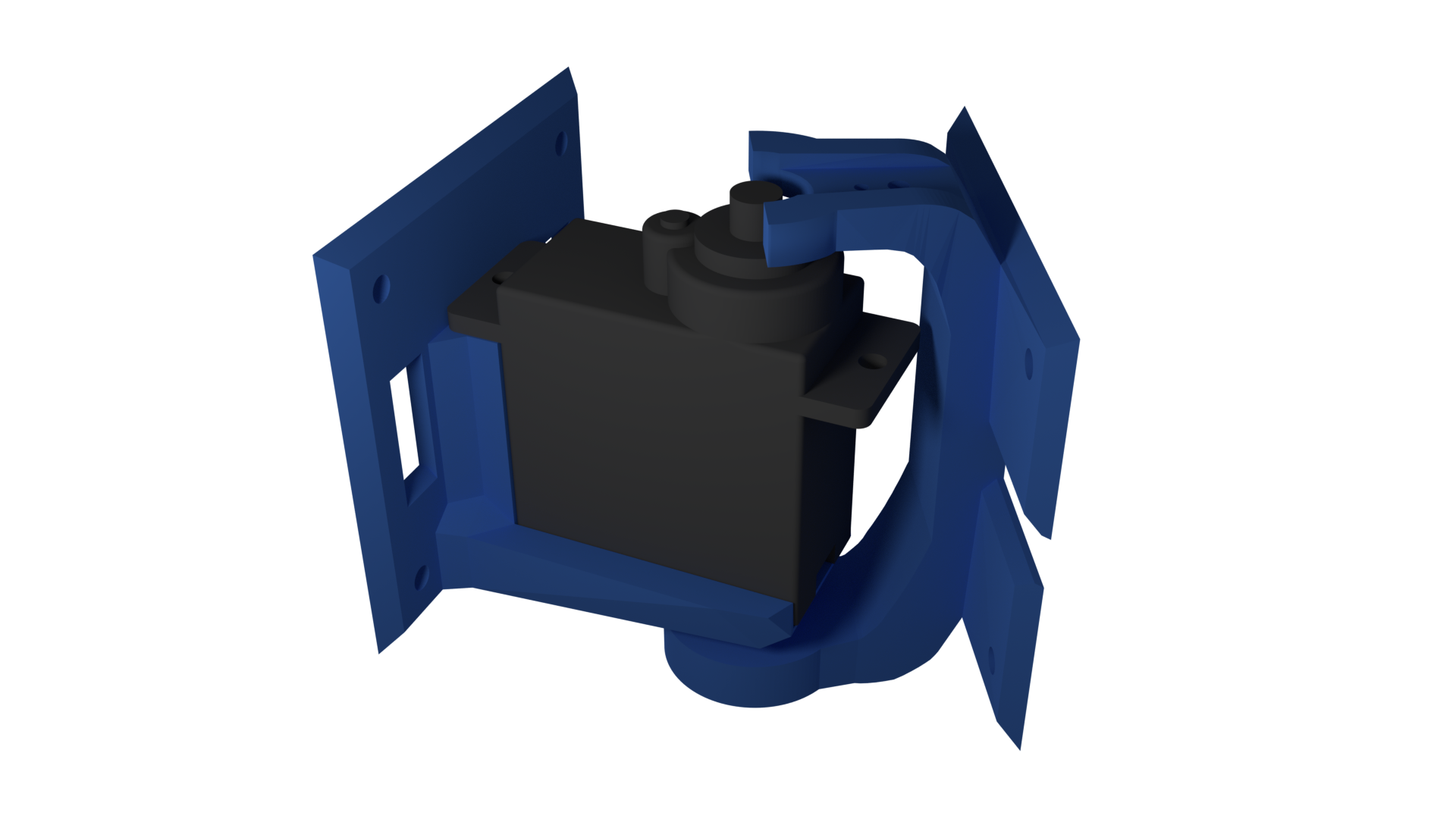}
        \caption{Joint module}
        \label{fig:robot_modules:joint}
    \end{subfigure}
    \caption{\small Robots are build using three types of modules. Starting with only one Core module, the robot grows by connecting Brick and Joint modules to the Core or other already connected modules.}
    \label{fig:robot_modules}
\end{figure}

The robots evolved in our system is a modular robotic framework based on RoboGen~\cite{auerbach_robogen_2014}.
Each robot is composed of three different types of modules:
one Core module (Figure~\ref{fig:robot_modules:core}), an arbitrary number of Brick modules (Figure~\ref{fig:robot_modules:brick}), and an arbitrary number of Joint modules (Figure~\ref{fig:robot_modules:joint}).
The Core module is unique for each robot and represents the robot ``head''  that, in the original physical incarnation~\cite{jelisavcic_real-world_2017}, contains the main logic board and the battery.
The Core module has four connection points where other modules can be attached.
Brick modules represent the ``backbone'' of the robot.
Only through Brick modules, the robot can take up arbitrary shapes. 
Actuation can only be achieved through the Joint modules, thus Joint modules are the only modules capable of changing the state of the robot in the environment.
Joint and Brick modules can be attached to any other module in two different ways, which differ from each other by $90\degree$ for the axis perpendicular to the attachment plane.
In \cite{de_carlo_comparing_2020}, we had already introduced this rotational attachment, but it applied only to Joint modules.
Allowing the Joint to be attached rotated allows the robot to evolve morphologies that have more variety in terms of degrees of freedom in terms of actuation.
In this work, we also introduce rotational attachment for Brick modules, which potentially allows robots that extend also vertically against gravity. 
In general, the design allows the inclusion of sensors, but for this study, we do not use any.

\subsection{Robot Brains}
\label{section:method:brain}

The controller of the robots is based on Central Pattern Generators (CPG) after  \cite{lan_directed_2018, lan_learning_2020}. 
Every joint in the body has a corresponding CPG node that consists of three neurons.
Two of these neurons (that we call $x$ and $y$ neurons) are coupled by two-directional connections, one from $x$ to $y$, and one from $y$ to $x$.
By definition, the weights of these connections have the same value but the opposite sign.
The remaining neuron in a CPG node provides the output signal to the servo motor driving the given joint.
The corresponding weight is set at $1.0$ in each joint, thus a CPG node can be configured by just one parameter regulating the connections between $x$ and $y$.

\begin{figure}[b]
    \centering
    \includegraphics[width=0.8\linewidth]{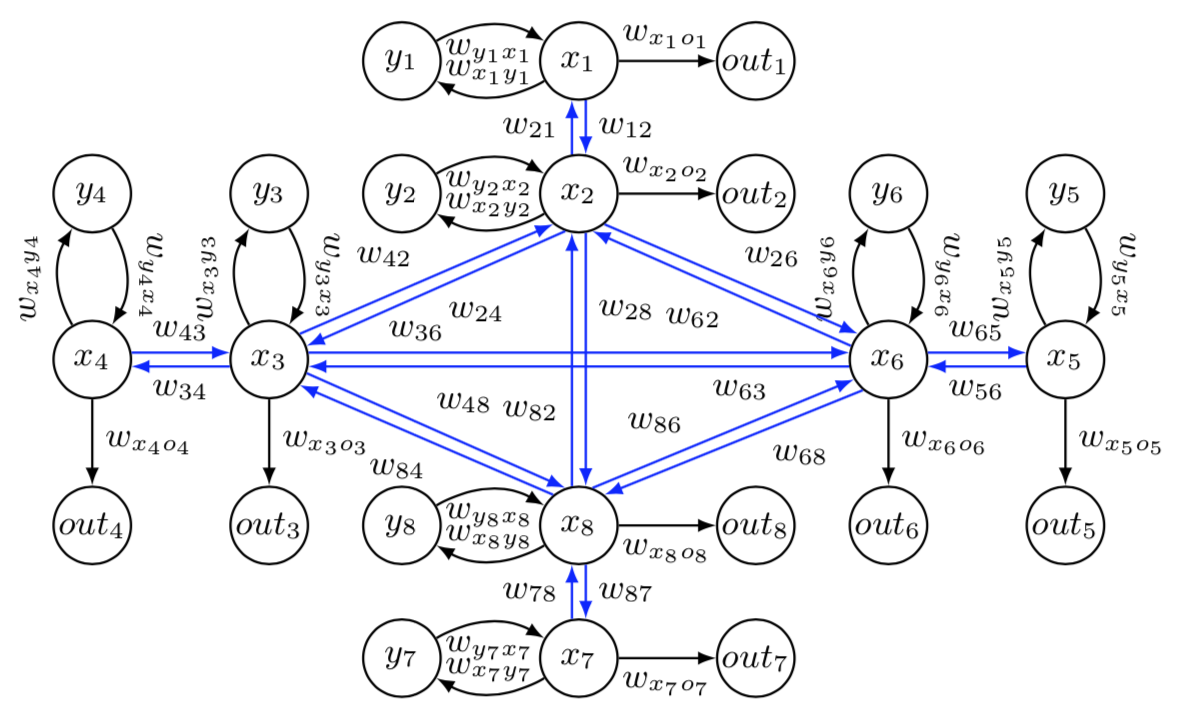}
    \caption{\small Brain diagram for a ''spider`` robot with eight joints positioned in a ''plus`` shape.}
    \label{fig:spider-brain}
\end{figure}

The overall controller architecture is a network with one CPG node for each joint and a connection between two such nodes if the corresponding joints are neighbors separated by no more than two empty cells (in the Manhattan sense) in the 3D Euclidean grid enclosing the robot.
Connected neighbor CPG nodes can synchronize the oscillations of their joints and induce global locomotion patterns.
The number of configurable parameters for a robot brain is thus $j+c$, where $j$ is the number of joints, and $c$ is the number of connections between joints; in Figure~\ref{fig:spider-brain} we show an example of how the nodes would be connected in a robot made of 8 joints configured as in the ''spider`` robot from \cite{jelisavcic_real-world_2017}.

\subsection{Evolution with tree-based representation}
For a direct encoding, we are using a tree-based representation, which in implementation is very similar to \cite{hupkes_revolve:_2018, jelisavcic_lamarckian_2019}.

In a tree-based representation, the genotype is a tree data structure in which each node represents a module of the robot.
Modules differentiate into three types: Core, Brick, and Joint modules, as represented in Figure~\ref{fig:robot_modules}.
The Core module is always the root of the tree and it can only be present once in the entire genotype. It can have four children.
The Brick module is attached on one side to its parent block and has three remaining slots available for child nodes.
The Joint module has only one remaining slot for a child module, therefore it does not allow any branching.
A Joint has three extra parameters that directly encode the oscillator parameters: frequency, offset, and amplitude.
In our tree-based representation, the brain development is limited to only decode parameters for the oscillators of the CPG network; i.e. all connection between oscillating nodes are not activated.

The robot module tree can be altered by one of the following mutation operators. The changes primarily revolve around changing the body: adding a random module, deleting a sub-tree, duplicating a sub-tree, or swapping a sub-tree. Alternatively, brain mutations are facilitated by mutating the joint oscillator parameters to achieve different activation patterns.

Parent robot trees can be recombined by inheriting sub-trees from the parents. Some checks and balances ensure that the recombined trees are valid, and do not exceed the maximum limit of modules.

\subsection{Evolution with L-system representation}

For an indirect encoding, we choose a system from our previous work \cite{de_carlo_comparing_2020}, which is composed of an Lindenmayer-system (L-system) \cite{lindenmayer_mathematical_1968} that describes body and brain structure and a component based on Hypercube-based Neuro Evolution of Augmenting Topologies (HyperNEAT) \cite{stanley_hypercube-based_2009,gauci_autonomous_2010} that encodes the weights of the CPG network.
These indirect encodings are capable of creating symmetrical growth structures and repetitions in our bodies.

L-Systems are parallel rewriting systems acting on a formal grammar.
The grammar is defined as a tuple $G=(V,w,R)$, where $V$ is the Alphabet, $w$ is the Axiom and $R$ is a set of Replacement Rules.
L-Systems start from the Axiom $w$, which is a sequence of symbols from the Alphabet.
To develop an L-System grammar, the Axiom is expanded into a longer sentence by replacing symbols using the Replacement Rules in $R$.
The replacement operation can be repeated multiple times on the sentence.

In this work, we adapted a system from~\cite{miras_impact_2019}, where each genotype is a grammar with always the same Axiom and Alphabet for all robots.
The Alphabet is made of the following symbols:
\begin{itemize}
    \item Robot modules: the Core, the Brick, a Vertical Joint, and a Horizontal Joint. 
    \item Mounting commands: $add\_left$, $add\_front$, and $add\_right$.
    Mounting commands must be followed by a module symbol otherwise they are ignored. When the final sentence is read, their role is to attach the following module symbol in the sentence to the module indicated by the cursor position, at a new position (left, front, or right) depending on the specific command.
    \item Moving commands: $move\_back$, $move\_right$, $move\_front$, and $move\_left$.
    The moving commands alter the position of the cursor.
\end{itemize}
The Replacement Rules of our L-System are a set of rules that replace any of the robot module symbols with a sequence of new symbols from the alphabet.
In other words, robot module symbols are both terminal and non-terminal symbols in our L-System.
Any other symbol is terminal, which means it cannot be replaced further.
The Axiom of our L-System is a sentence made of a single symbol: the Core block.
Once the L-System grammar develops the Axiom into a sentence, the sentence is used as a sequence of instructions that describe how to build the robot.

The system we used here differs from the one we inspired upon, by the addition of an additional constraint on the morphology, i.e. we do not allow a joint to be attached to another joint.
This extra addition showed in previous experiments~\cite{de_carlo_comparing_2020} that increase the chances for more complex robots to appear and we decided to use it here to increase the chances for interesting morphologies in both experimental configurations.
We also improved the Alphabet with the introduction of a new rotated Brick module, which allows for the morphologies to develop in three dimensions.

The brain structure is defined by the body; each joint creates a corresponding CPG oscillator node and connections are made using the rules already explained in subsection~\ref{section:method:brain}.
When all connections are defined, each CPG node is positioned on a substrate space with $x,y,z,w$ coordinates and each connection weight value is queried from a CPPN \cite{Stanley2007}, as defined by the HyperNEAT algorithm.
In the substrate space, $x$, $y$, and $z$ determine the position of the CPG's node corresponding joint, while the $w$ axis determines the front and back neurons in an oscillating CPG node.

The mutation and crossover operators are defined by their individual components:
for the L-system component, we use the operators defined in~\cite{miras_effects_2019}.
For the HyperNEAT component, we use the operators as defined by HyperNEAT, with the exclusion of the species, i.e. genomes are not divided into species and crossover is possible for each pair of genome in the population.

\section{Methodology}
\subsection{Heritability}
In an evolutionary system the phenotypic variation ($V_P$) of a population of individuals is an expression of genetic variation ($V_G$) and environmental factors ($E$).
\begin{equation}
V_P = V_G + E
\end{equation}
The genetic variation can be further subdivided into three major components: additive genetic variation ($V_A$), non additive genetic variation caused by epistatic genes ($V_{NA}$) and effects of random mutations ($M$).
\begin{equation}
V_G = V_A + V_{NA} + M
\label{eq:genetic_variation}
\end{equation}
As defined in \cite{wray2008estimating,griffiths_quantifying_2000}, heritability measures the contribution of genes to phenotypic traits.
Each phenotypic trait has a different value of heritability.
Heritability can be defined as broad-sense heritability ($H^2$) or narrow-sense heritability ($h^2$).
Broad-sense heritability is the proportion of phenotypic variation that is created by the genetic variation.
Narrow-sense heritability is only the proportion of genetic variation that is generated by additive genetic values, not including any effect of dominance or epistasis.
\begin{equation}
H^2 = \frac{V_G}{V_P}
\hspace{1cm}
h^2 = \frac{V_A}{V_P}
\end{equation}
Heritability is also an important component of the ``response to selection'' ($R$), a value that can be predicted as the product of narrow-sense heritability and selection differential ($S$) \cite{lynch1998genetics}:
\begin{equation}
    R = h^2 \cdot S    
\end{equation}

The value of heritability can be calculated from its theoretical formula, but this requires a deep mathematical understanding of our genotype model.
However, if we are only interested in the additive genetic material, an estimate of narrow-sense heritability can easily be derived from population measurements by linearly regressing the average trait value of the offspring against the parental phenotype.
An approximation to a linear model is possible because the additive genetic code has a linear response to the resulting phenotype, in contrast to epistatic genetic code which has a much more unpredictable effect on the phenotype.
The value for the slope of the linear regression is our numerical estimation for heritability.
The value for heritability can vary between $1.0$ and $0.0$, where $h^2 = 1.0$ is a $45\degree$ linear regression, representing a perfect match between parents' average trait and the offspring's trait.
A value of $h^2 = 0.0$ instead represents a scenario where the offspring's trait is completely unpredictable given the parents' traits.

In biological systems, heritability can isolate those phenotypic features that are expression of genetic material from features that are influenced by the environment.
Estimated heritability for life history traits and behaviour are typically low to medium (ranging up to $0.30$~\cite{dochtermann2019heritability}), whereas morphological traits are often found to have higher heritability (average $h^2 = 0.46$~\cite{mousseau1987natural}).

In evolutionary robotics, the influence of the environment over the development of the individuals is usually very limited, excluding a few exceptions \cite{auerbach_environmental_2014,miras_environmental_2020}.
In our case the environment is a flat terrain, therefore its influence is completely absent. 
The implication is that the phenotypic variation in this system is only an expression of genotypic variation.
Through linear regression we can estimate the narrow-sense heritability, which is only an expression of additive genetic variation.
The rest of the phenotypic variation can only be an expression of epistatic gene interaction and mutation (Eq.~\ref{eq:genetic_variation}).


\subsection{Robot Traits}
\label{section:setup:measures}

To estimate an overall heritability of our system, we chose a wide variety of phenotypic traits that are representative of different aspects of our robot.
This work is interested in the overall evolution of modular robots, but the approach is not limited to any particular number or type of traits.
The same study can be repeated on any trait, e.g. it would be interesting to study the heritability of ``the number of feet in a robot'' and what parameters increase the transmission of the trait to the offspring.

We recorded a set of many traits derived from the descriptors found in~\cite{miras_environmental_2020}.
From the many traits available, we sampled only a significant few that we found to be orthogonal to each other in previous work \cite{carlo_influences_2020}: some traits that measure the morphological aspect of the robots and some that measure the behavioural aspect.

The \textbf{Morphological} traits give us insight on how the robot shapes evolve. 
In this work we used:
\begin{itemize}
    \item \emph{Proportion}: considering the 2D bounding box that encompasses the robot when viewed from above, this trait is the ratio between the two sides of this rectangle.
    \item \emph{Size}: the number of modules in the body.
    \item \emph{Number of Limbs}: considering the robot as a tree of modules, it is the number of leaf modules. The value is normalized per robot by the number of all possible limbs available.
    \item \emph{Coverage}: considering the 3D bounding box that encompasses the robot, this trait is the ratio between the area that is occupied by modules and the total area of the rectangle.
\end{itemize}

The \textbf{Behavioural} traits are very important because they give insight in the complex relationship of body and brain.
In this work we used:
\label{section:behavioural_traits_displacement_speed}
\begin{itemize}

    
    \item \emph{Speed}: Describes the average robot speed ($cm/s$), and is calculated as if the robot took the shortest path from the start position $s_0$ to the end position $s_t$, and is defined with Eq.~\ref{eq:displacement_speed}. 
    \label{section:behavioural_traits_displacement_speed}
    \begin{equation}
        v_{disp} = \frac{s_{T} - s_{t_0}}{\Delta{t}}
        \label{eq:displacement_speed} 
    \end{equation}
    
    \item \emph{Balance}: We use the rotation of the head in the $x$--$y$ plane to define the balance of the robot. 
    We describe the rotation of the robot with three dimensions: roll $\phi$, pitch $\theta$, and yaw $\psi$. 
    Thus, we consider the pitch and roll of the robot head, expressed between $0\degree$ and $180\degree$ (because we are not interested in whether the rotation is clockwise or anti-clockwise).
    Perfect balance corresponds to $\theta = \phi = 0\degree$, so that the higher balance, the less rotated the head is. Formally, balance is defined by Eq.~\ref{eq:balance}.
    \begin{equation}
        b = 1 - \frac{\sum^T_{t=1}{|\phi_t| + |\theta_t|}}{180 \cdot 2 \cdot \Delta{t}}
        \label{eq:balance} 
    \end{equation}

\end{itemize}
\section{Setup}
For both encodings we use the same evolutionary algorithm with a generational population update scheme, that is, an evolutionary algorithm where consecutive populations are non-overlapping. This means that survivor selection is trivial: no members of population $P_n$ survive, the subsequent generation $P_{n+1}$ consists of offspring of the current one. As for parent selection, we use the tournament selection mechanism with a tournament size of two individuals. This represents a low selection pressure.  We run this algorithm with a population size of 100 individuals for 50 generations, amounting to a total of 5000 evaluations as the computational budget for optimizing the robots' makeup. For both encodings, fitness evaluations are done by placing the given robot on a flat surface and running it for 30 seconds.
We evolved the robots for movement, using the \emph{speed} behavioral trait as the value for fitness.
We adjusted the mutation rates for the evolutionary run to be quite high, with a probability of $0.59$ of having at least a body mutation for the tree-based representation and we used the same probability ($0.59$) for the mutation chance on the L-System grammar.

\section{Results Analysis}
\begin{table}[]
    \centering
    \begin{tabular}{rcc}
        \toprule
         & \multicolumn{2}{c}{Heritability} \\
         \multicolumn{1}{c}{Trait}
                    & Tree-based & L-System  \\
        \midrule
        Speed       & 0.74       & 0.35 \\
        Balance     & 0.77       & 0.37 \\
        Proportion  & 0.65       & 0.41 \\
        Size        & 0.73       & 0.47 \\
        N. of limbs & 0.83       & 0.67 \\
        \bottomrule
    \end{tabular}
    \caption{\small Estimated values of heritability for each phenotypic trait divided}
    \label{tab:heritability_all_values}
\end{table}
\newcommand{\speedfigsize}{.38}
\begin{figure*}
    \centering
    \begin{subfigure}[t]{\speedfigsize\linewidth}
        \centering
        \includegraphics[width=.9\linewidth]{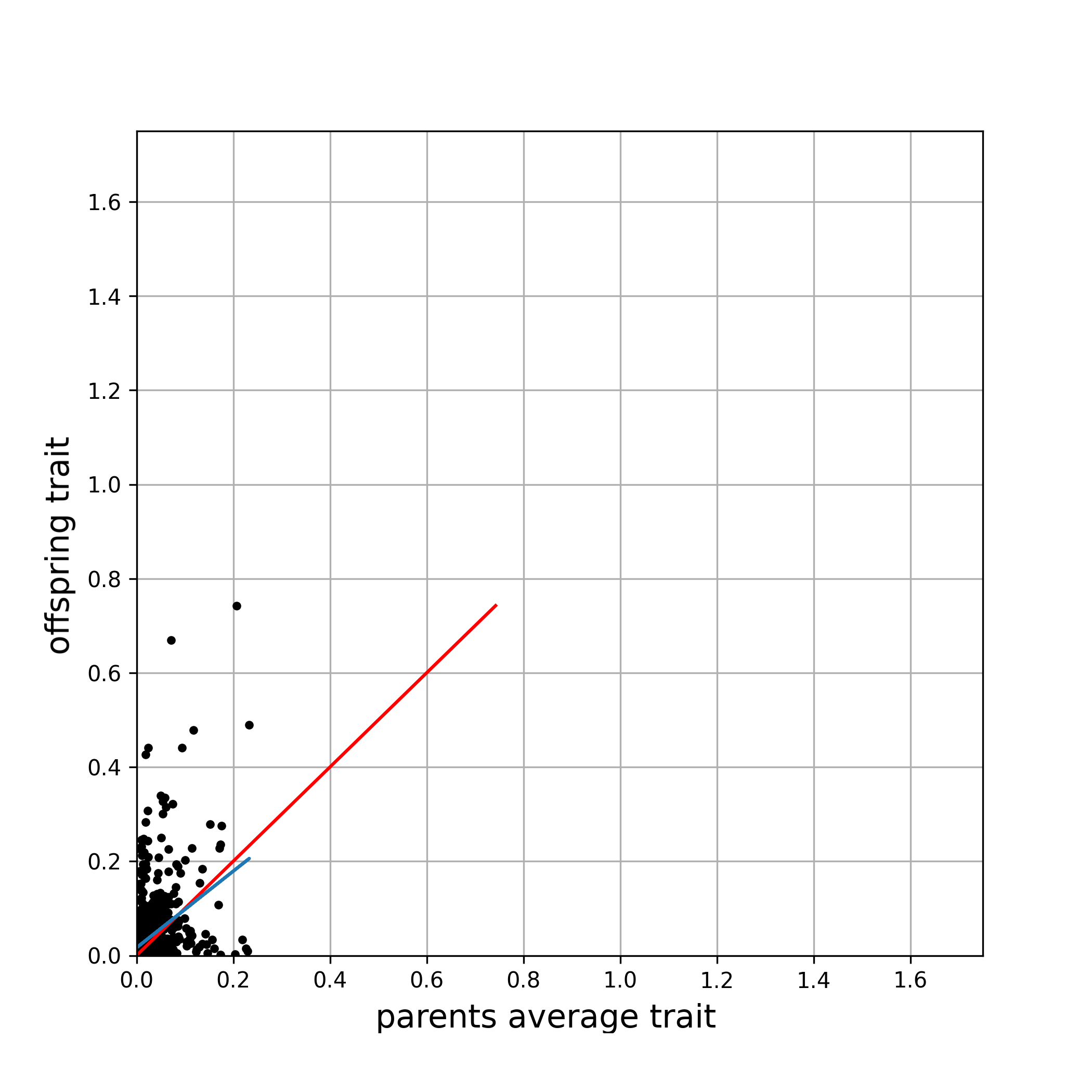}
        \caption{Heritability of tree-based encoding for the speed trait. $H^2 = 0.74$. Blue line is the resulting linear regression (slope is heritability). Red line is a perfect heritability reference. We can observe that even if points are more concentrated, they have an higher heritabiltity compared to the tree-based encoding.}
        \label{fig:speed:scatter_tree}
    \end{subfigure}
    \hspace{.5cm}
    \begin{subfigure}[t]{\speedfigsize\linewidth}
        \centering
        \includegraphics[width=.9\linewidth]{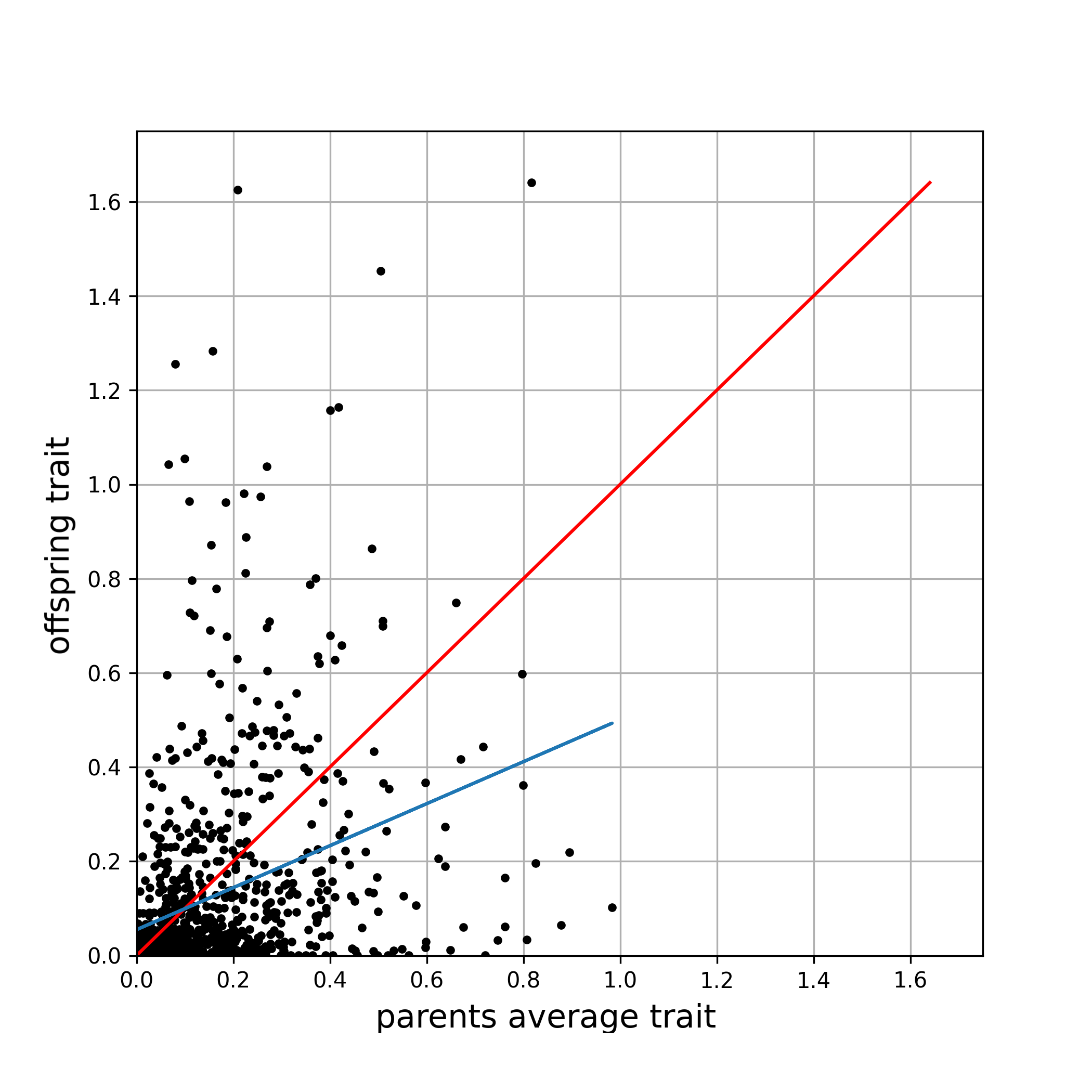}
        \caption{Heritability of L-System encoding for the speed trait. $H^2 = 0.35$. Blue line is the resulting linear regression (slope is heritability). Red line is a perfect heritability reference. We can observe that even if points are more scattered, they have a lower heritabiltity compared to the tree-based encoding. }
        \label{fig:speed:scatter_lsystem}
    \end{subfigure}
    \begin{subfigure}[b]{\speedfigsize\linewidth}
        \centering
        \includegraphics[width=.9\linewidth]{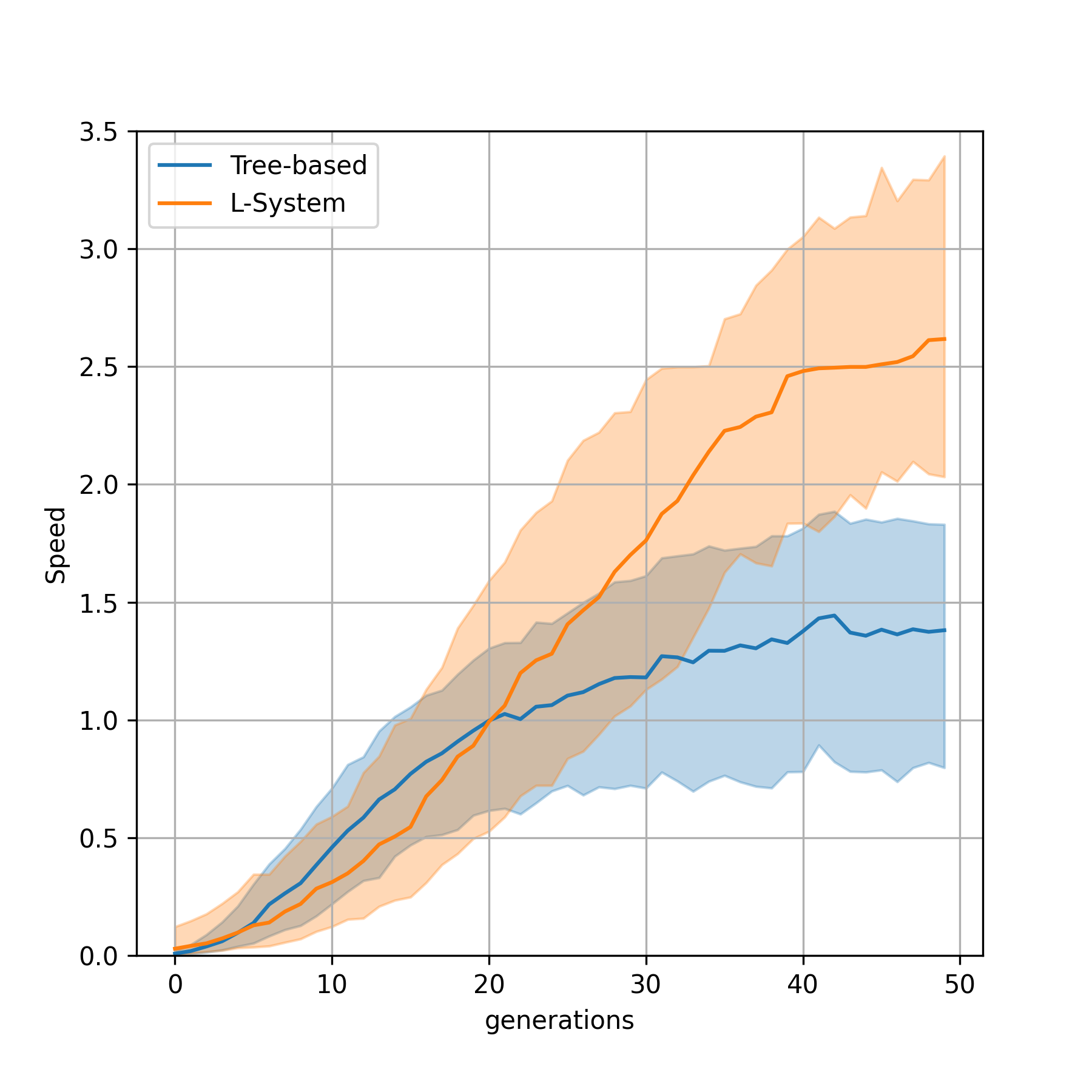}
        \caption{Median speed per generation}
        \label{fig:speed:median}
    \end{subfigure}
    \hspace{.5cm}
    \begin{subfigure}[b]{\speedfigsize\linewidth}
        \centering
        \includegraphics[width=.9\linewidth]{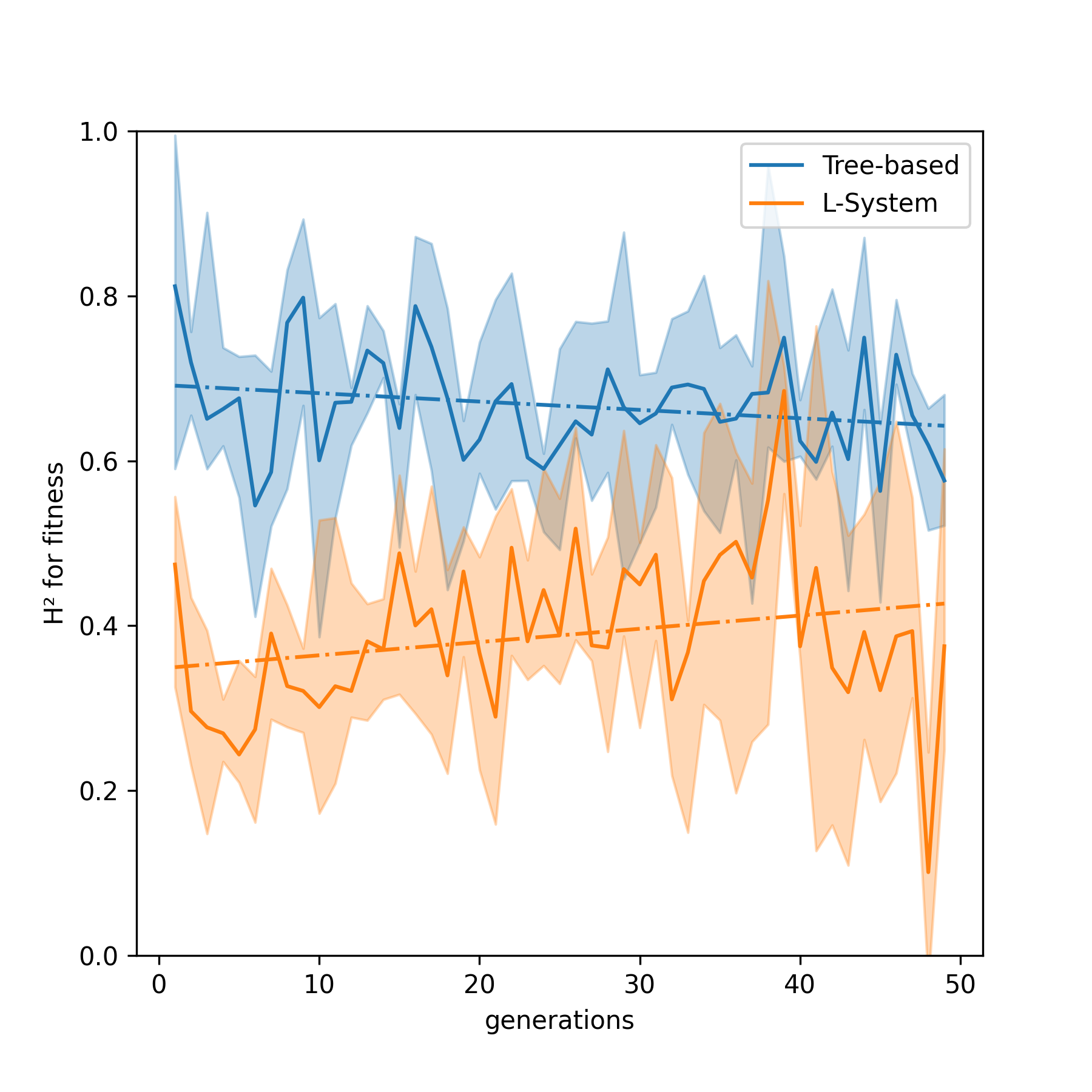}
        \caption{Heritability per generation}
        \label{fig:speed:heritability_generation}
    \end{subfigure}
    \begin{subfigure}[b]{\speedfigsize\linewidth}
        \centering
        \includegraphics[width=.9\linewidth]{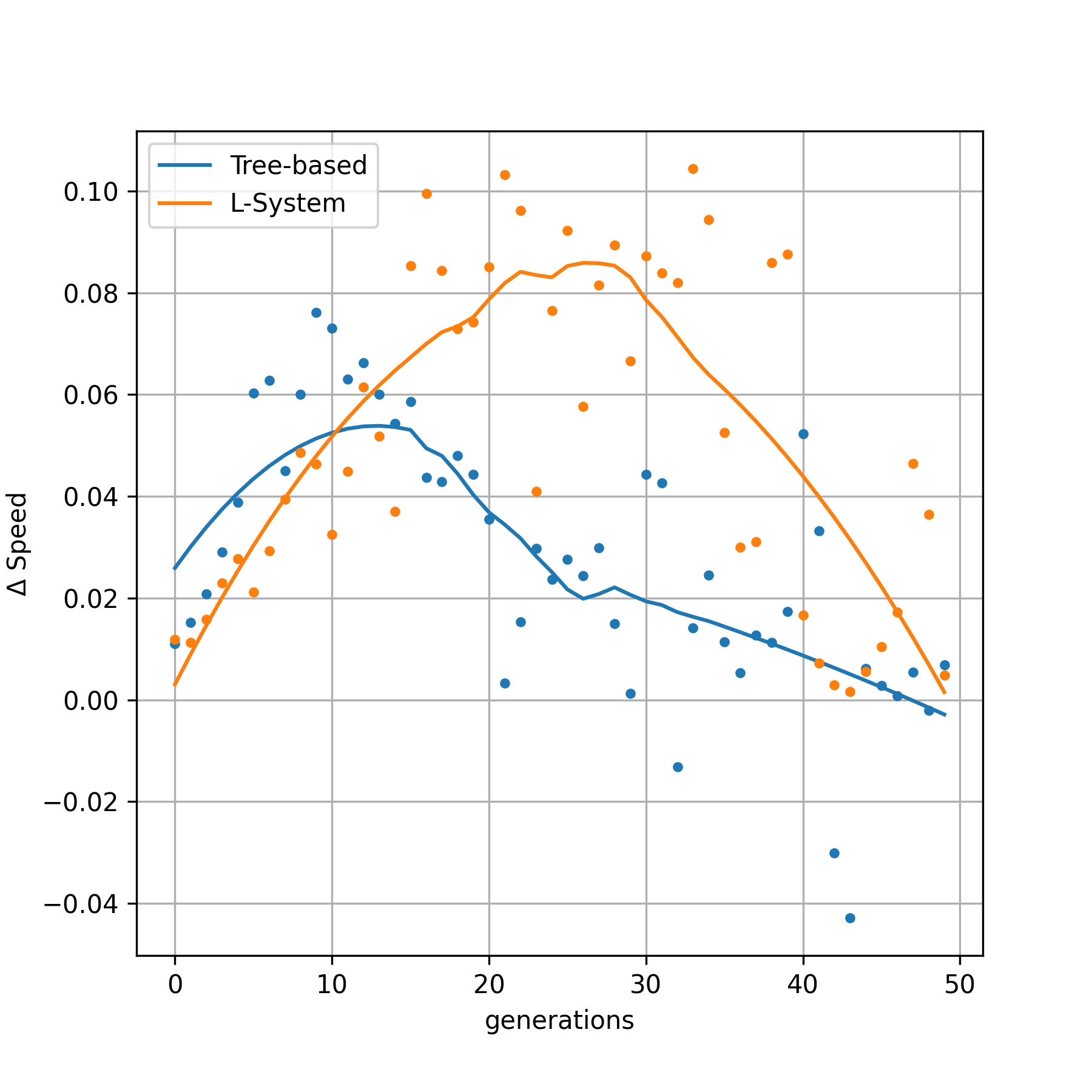}
        \caption{Increment of the median speed per generation}
        \label{fig:speed:derivate}
    \end{subfigure}
    \hspace{.5cm}
    \begin{subfigure}[b]{\speedfigsize\linewidth}
        \centering
        \includegraphics[width=.8\linewidth]{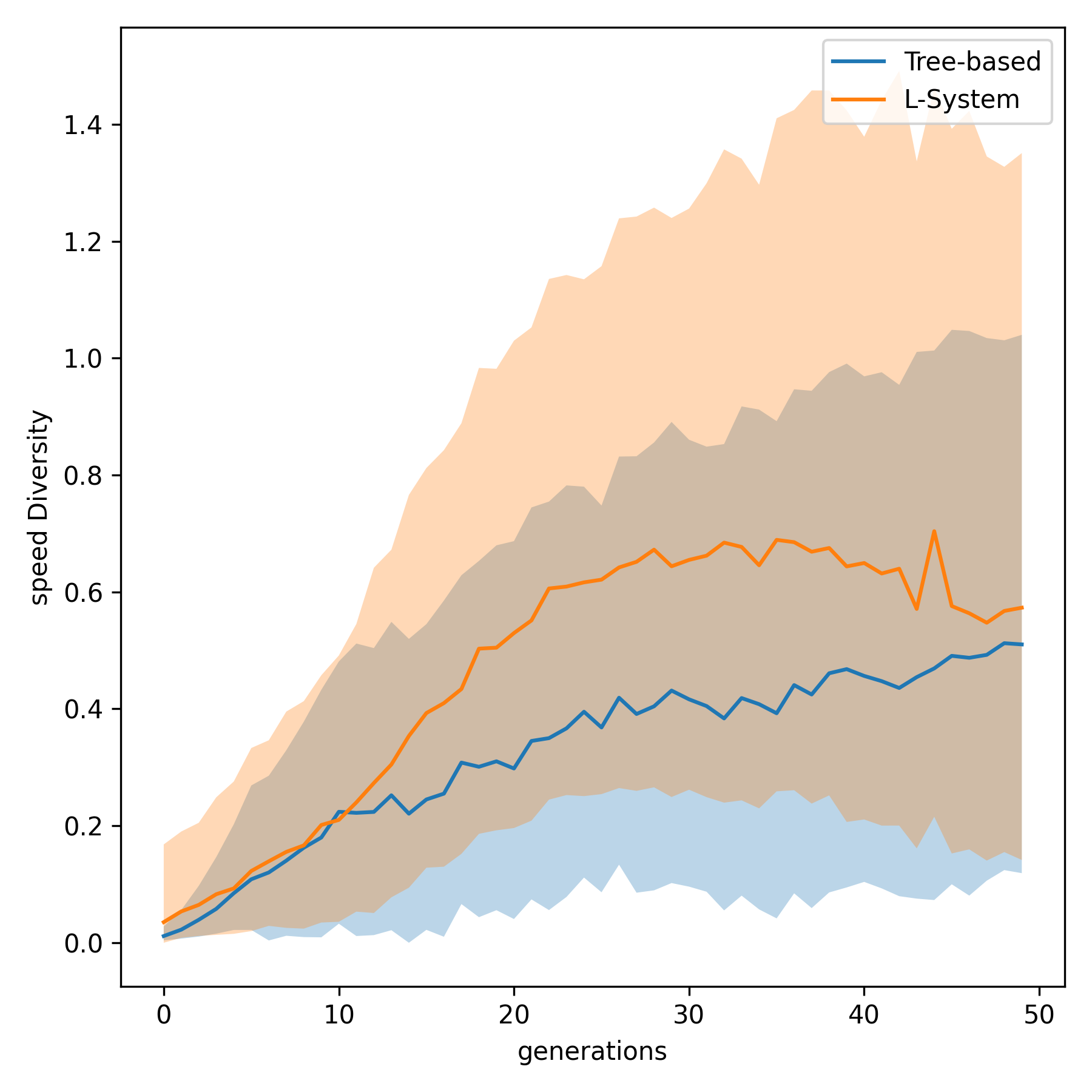}
        \caption{Median diversity of speed per generation}
        \label{fig:speed:diversity}
    \end{subfigure}
    \vspace{.3cm}
    \caption{Analysis of behaviour. Here we consider the evolution of the speed trait, how fast it changes per generation, speed heritability at the first generation and how it changes during evolution, and the diversity of speed present across the population.}
    \label{fig:speed}
\end{figure*}

\newcommand{\limbsfigsize}{.38}

\begin{figure*}
    \centering
    \begin{subfigure}[b]{\limbsfigsize\linewidth}
        \centering
        \includegraphics[width=.9\linewidth]{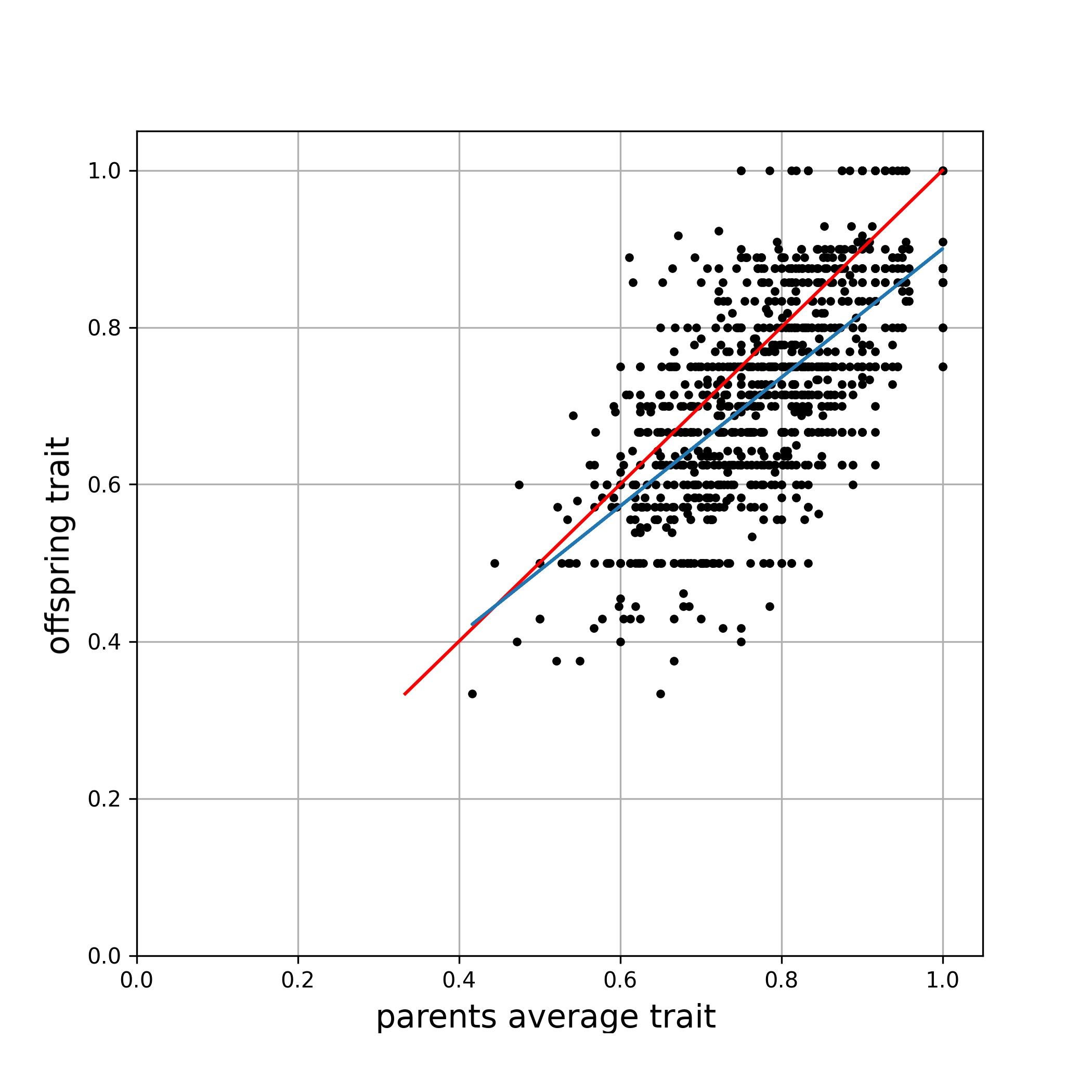}
        \caption{Heritability of tree-based encoding for the ``number of limbs'' trait. $H^2 = 0.83$. Blue line is the resulting linear regression (slope is heritability). Red line is a perfect heritability reference. We can observe that points are well clustered around the $45\degree$ line, resulting in a pretty high value of heritability.}
        \label{fig:limbs:scatter_tree}
    \end{subfigure}
    \hspace{.5cm}
    \begin{subfigure}[b]{\limbsfigsize\linewidth}
        \centering
        \includegraphics[width=.9\linewidth]{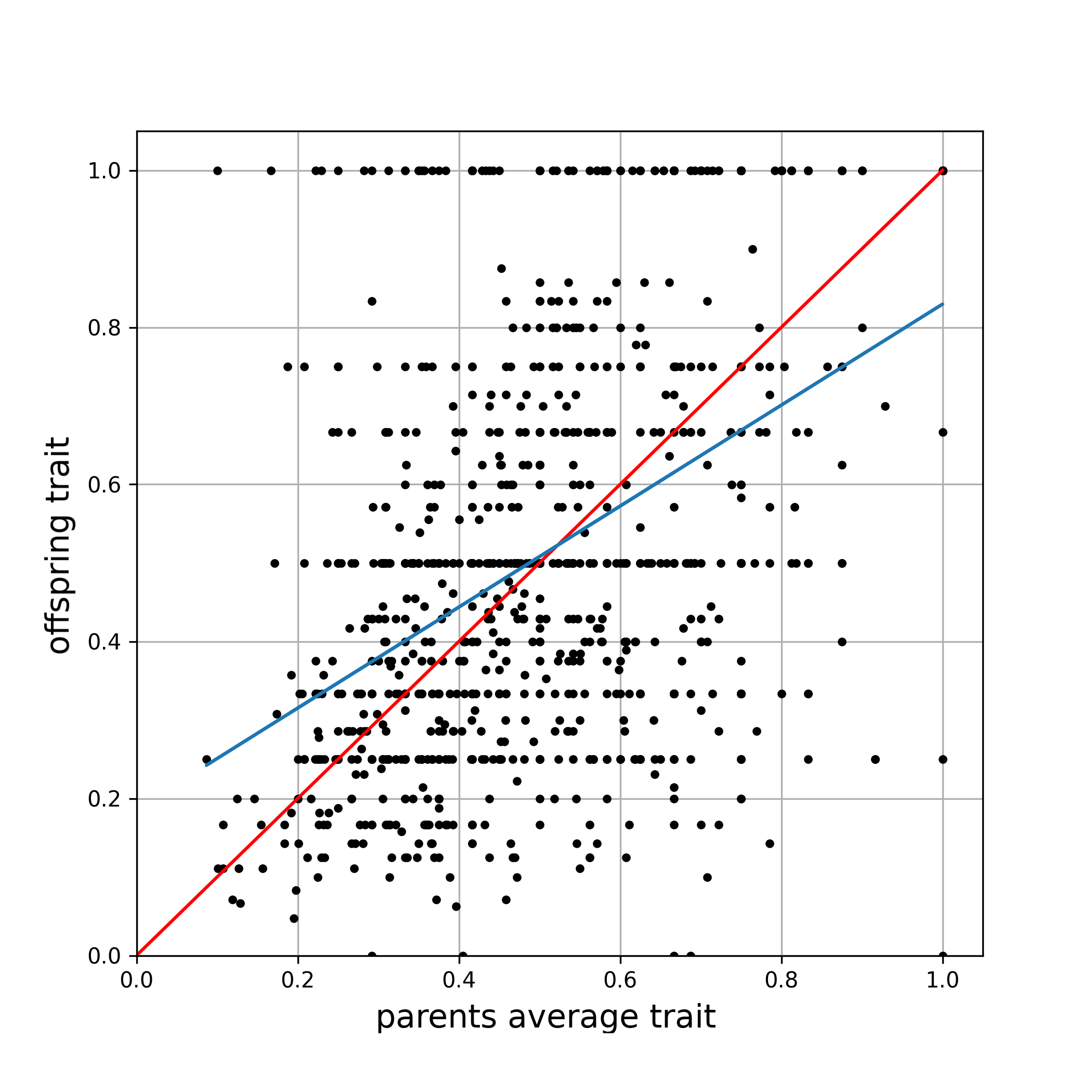}
        \caption{Heritability of L-System encoding for the ``number of limbs'' trait. $H^2 = 0.67$. Blue line is the resulting linear regression (slope is heritability). Red line is a perfect heritability reference. We can observe that points are scattered all around the plot, resulting in a lower value of heritability.}
        \label{fig:limbs:scatter_lsystem}
    \end{subfigure}
    \begin{subfigure}[b]{\limbsfigsize\linewidth}
        \centering
        \includegraphics[width=.9\linewidth]{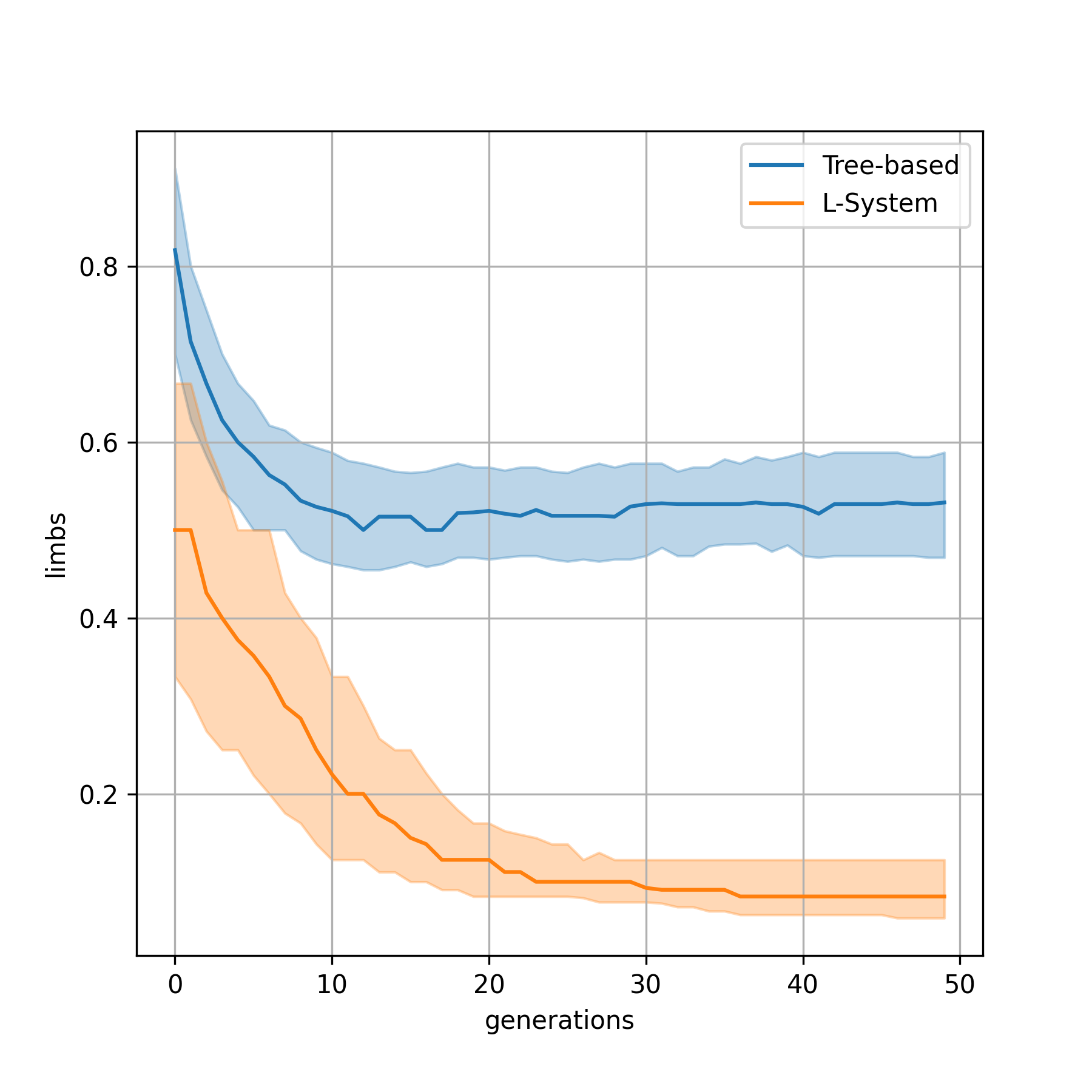}
        \caption{Median number of limbs per generation}
    \end{subfigure}
    \hspace{.5cm}
    \begin{subfigure}[b]{\limbsfigsize\linewidth}
        \centering
        \includegraphics[width=.9\linewidth]{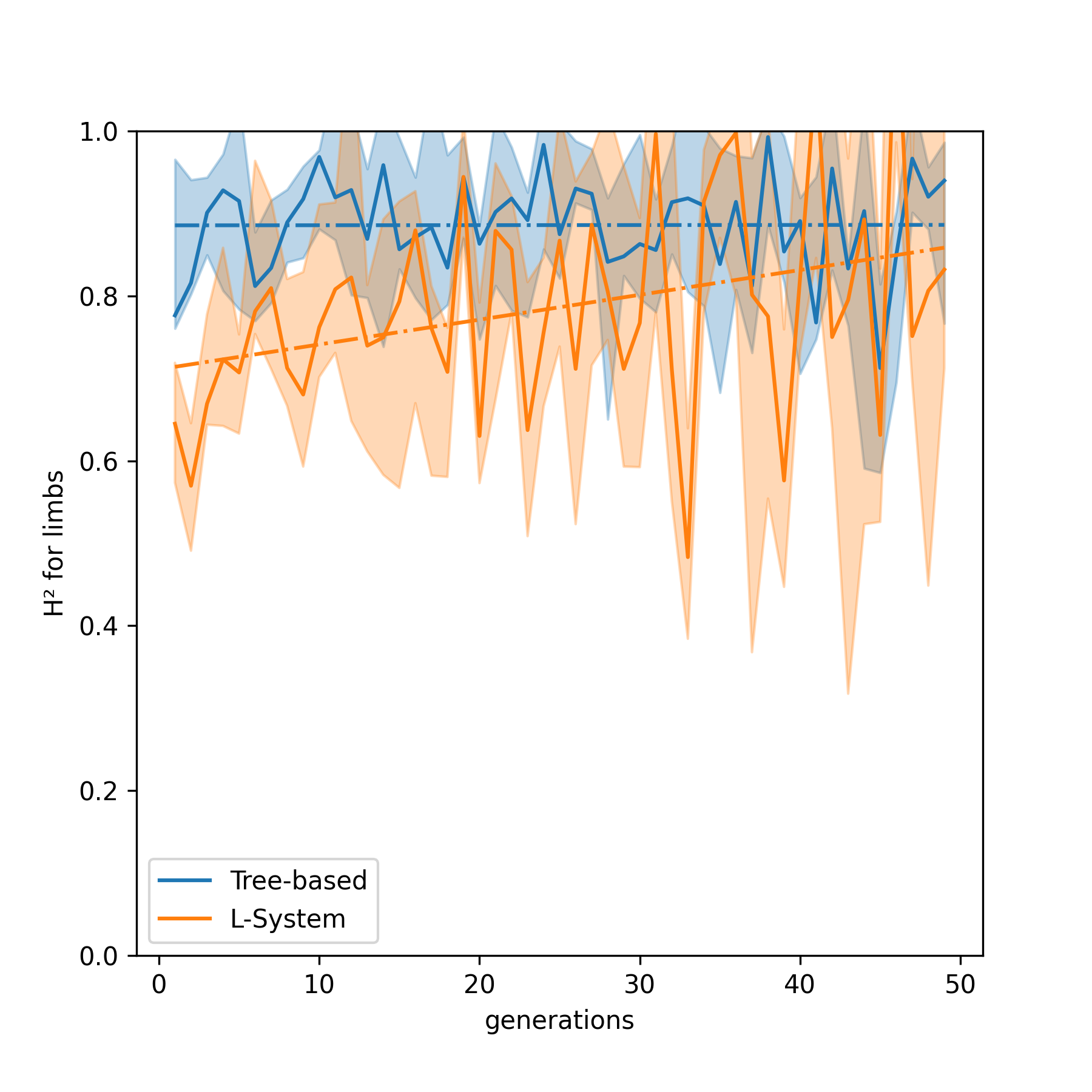}
        \caption{Heritability per generation}
        \label{fig:limbs:heritability_generation}
    \end{subfigure}
    \begin{subfigure}[b]{\limbsfigsize\linewidth}
        \centering
        \includegraphics[width=.9\linewidth]{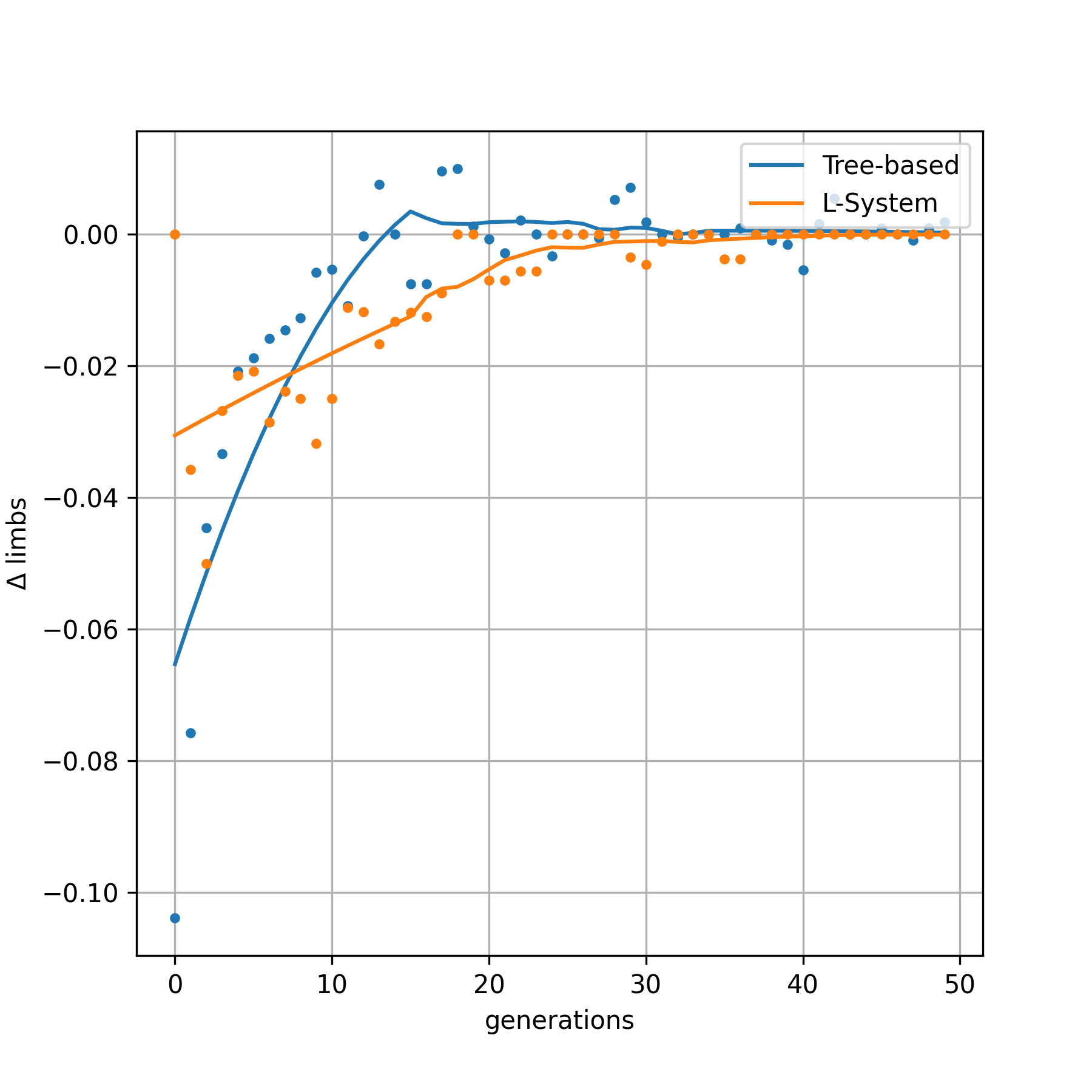}
        \caption{Increment of the median number of limbs per generation}
        \label{fig:limbs:derivate}
    \end{subfigure}
    \hspace{.5cm}
    \begin{subfigure}[b]{\limbsfigsize\linewidth}
        \centering
        \includegraphics[width=.8\linewidth]{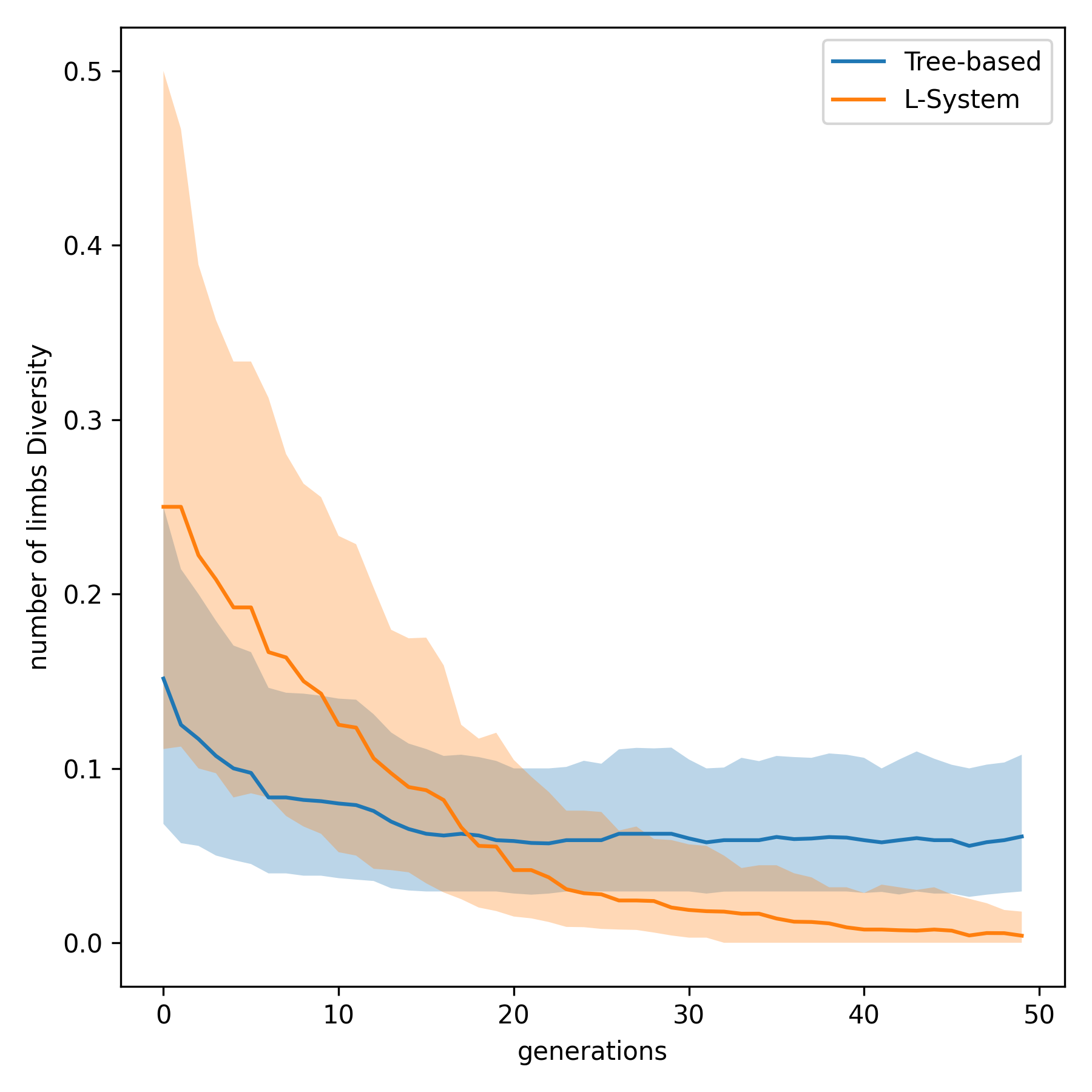}
        \caption{Median diversity of number of limbs per generation}
    \end{subfigure}
    \vspace{.3cm}
    \caption{Analysis of morphologies. Here we consider the evolution of the ``number of limbs'' trait, how fast it changes per generation, the trait's heritability at the first generation and how it changes during evolution, and the diversity of the ``number of limbs'' present across the population. The value for the ``number of limbs'' is normalized to the maximum possible number (of limbs) for each robot.}
    \label{fig:limbs}
\end{figure*}

\newcommand{\scattersize}{0.32}
\begin{figure*}
    \centering
    \begin{subfigure}[b]{\scattersize\linewidth}
        \includegraphics[width=\linewidth]{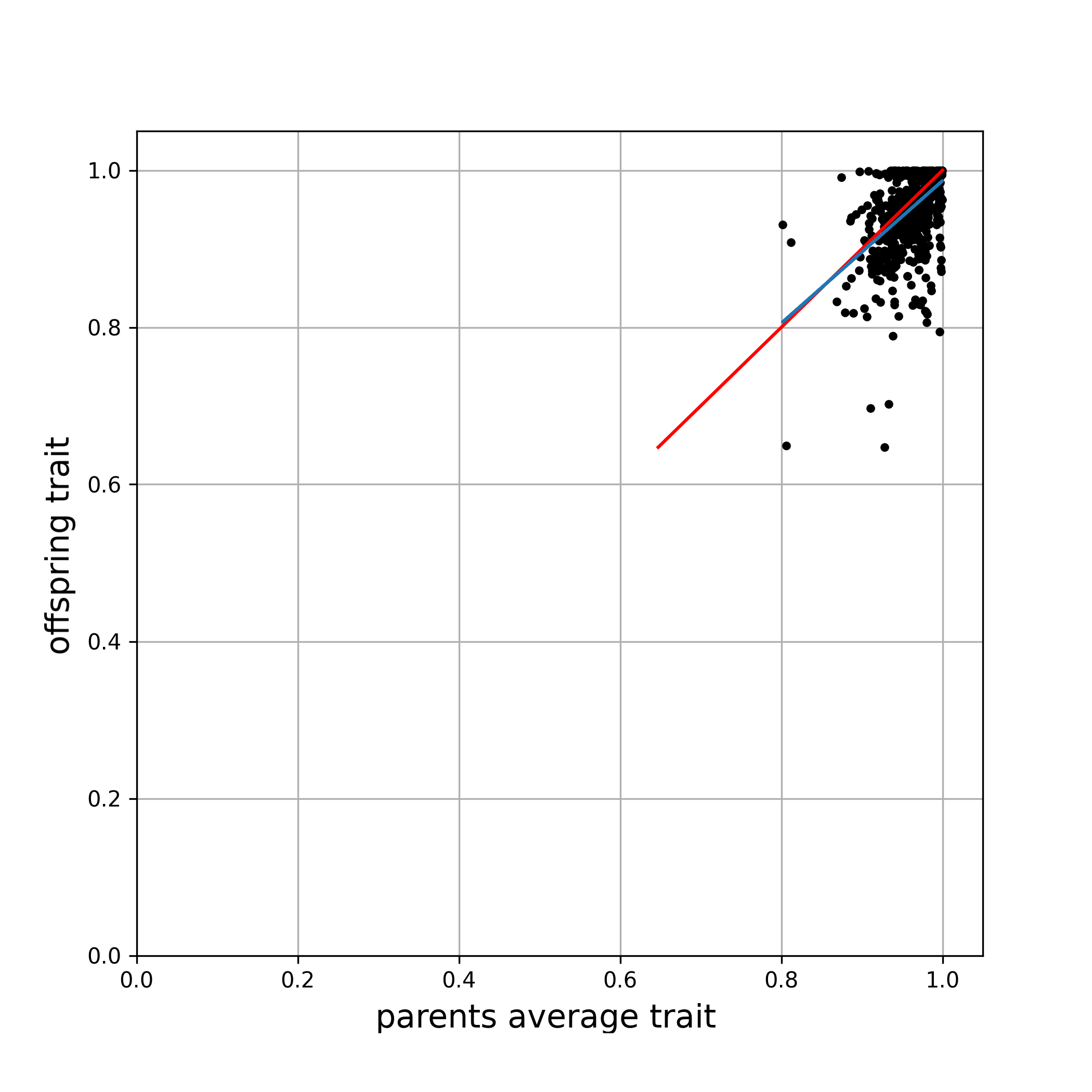}
        \caption{Balance: Tree-based}
    \end{subfigure}
    \begin{subfigure}[b]{\scattersize\linewidth}
        \includegraphics[width=\linewidth]{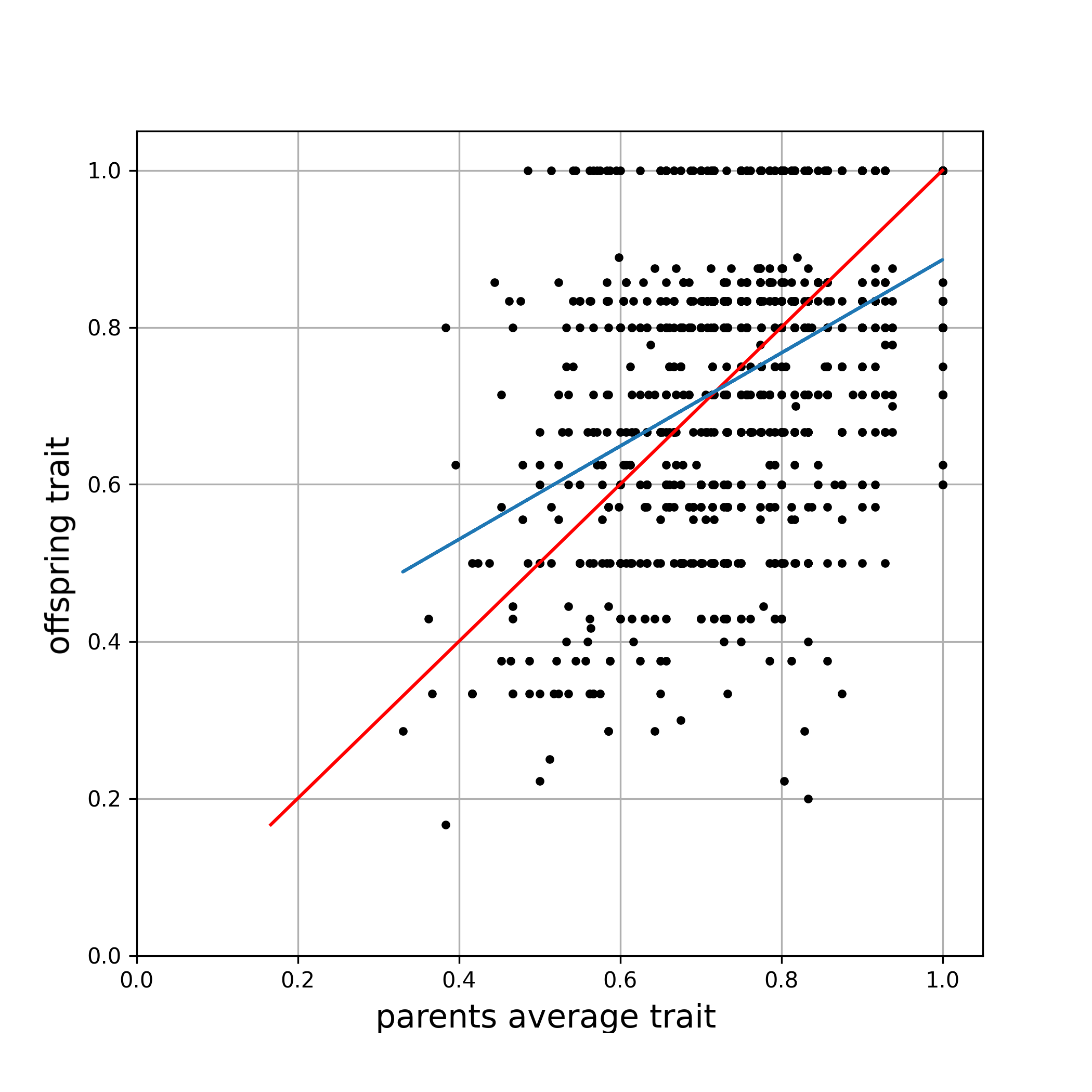}
        \caption{Proportion: Tree-based}
    \end{subfigure}
    \begin{subfigure}[b]{\scattersize\linewidth}
        \includegraphics[width=\linewidth]{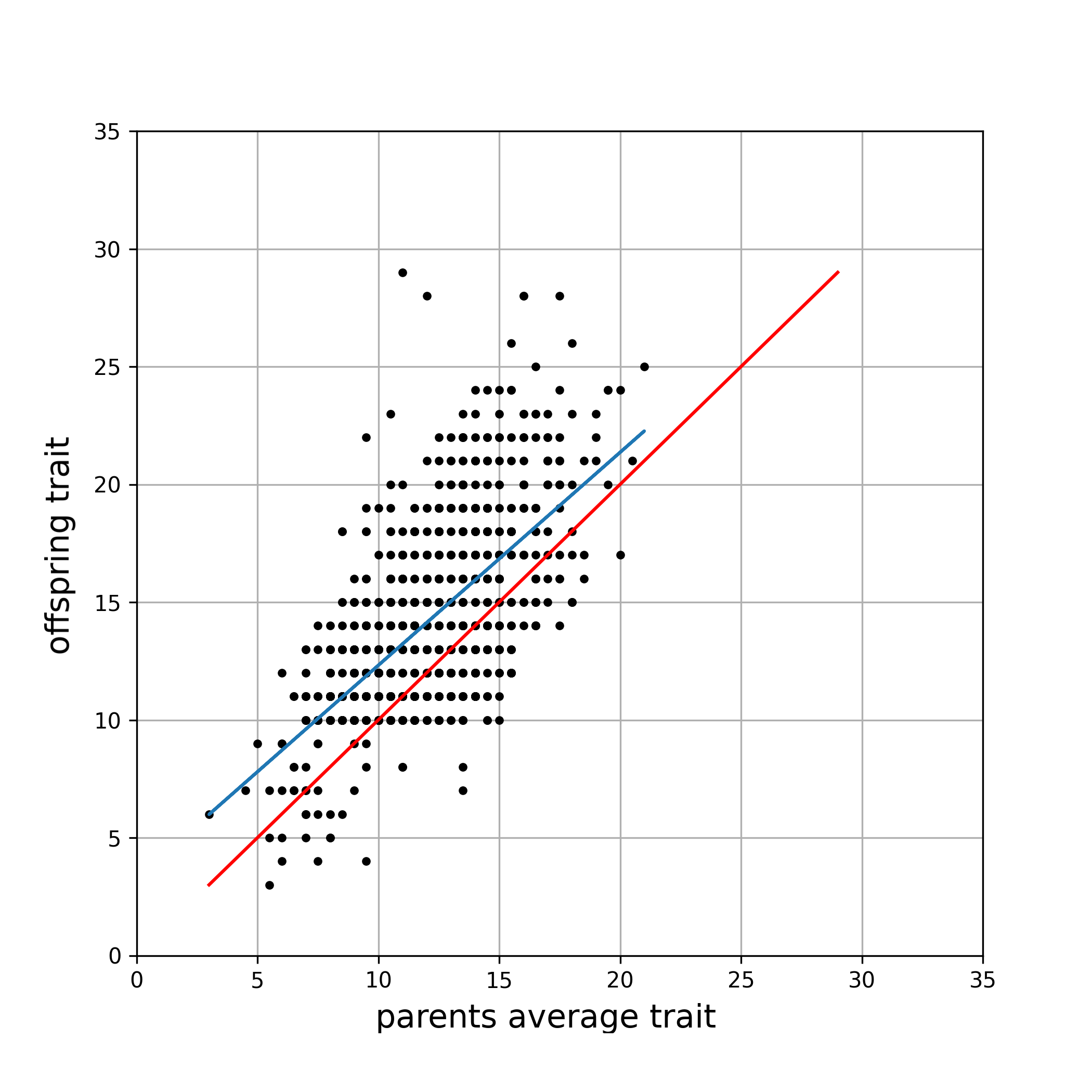}
        \caption{Size: Tree-based}
    \end{subfigure}
    \begin{subfigure}[b]{\scattersize\linewidth}
        \includegraphics[width=\linewidth]{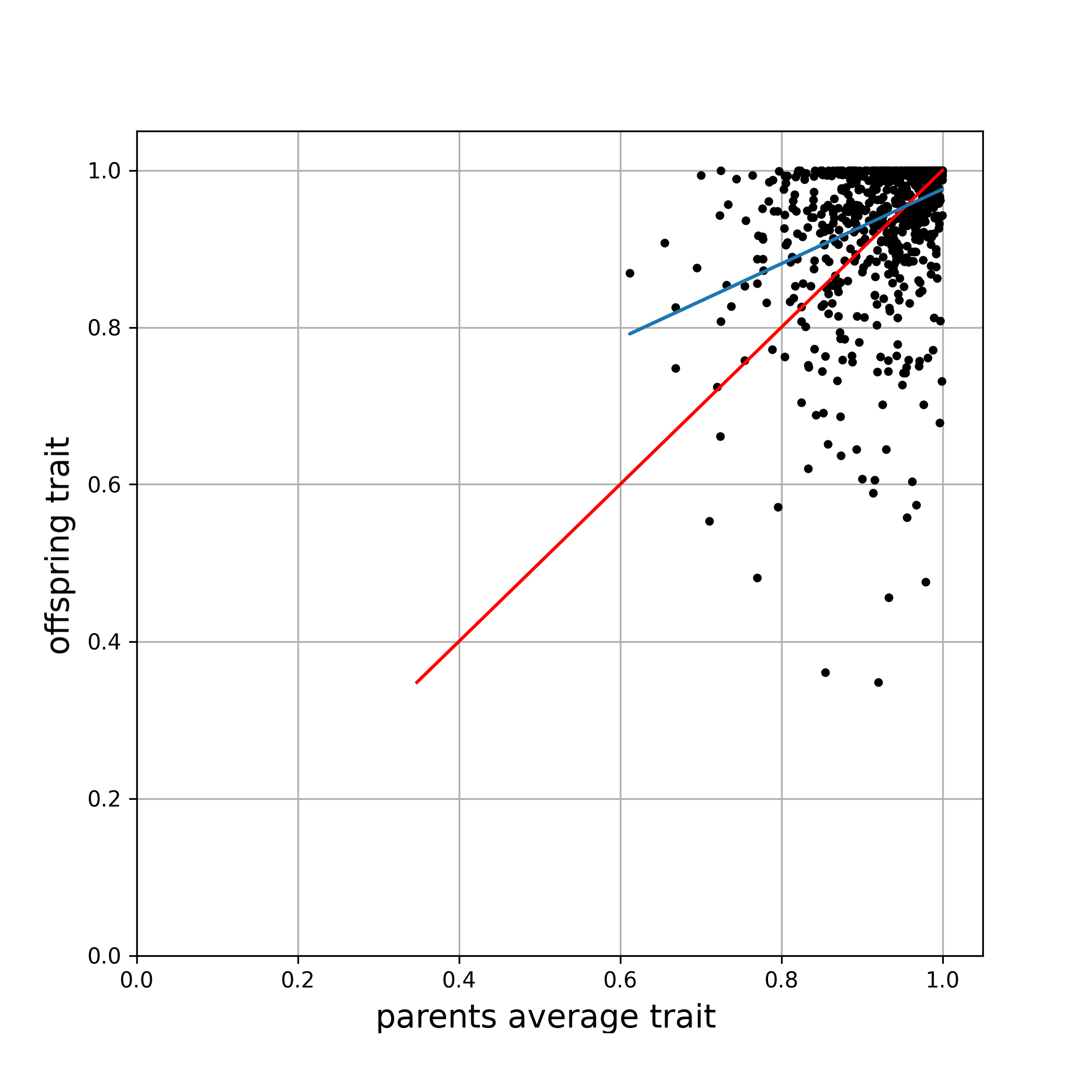}
        \caption{Balance: L-system}
    \end{subfigure}
    \begin{subfigure}[b]{\scattersize\linewidth}
        \includegraphics[width=\linewidth]{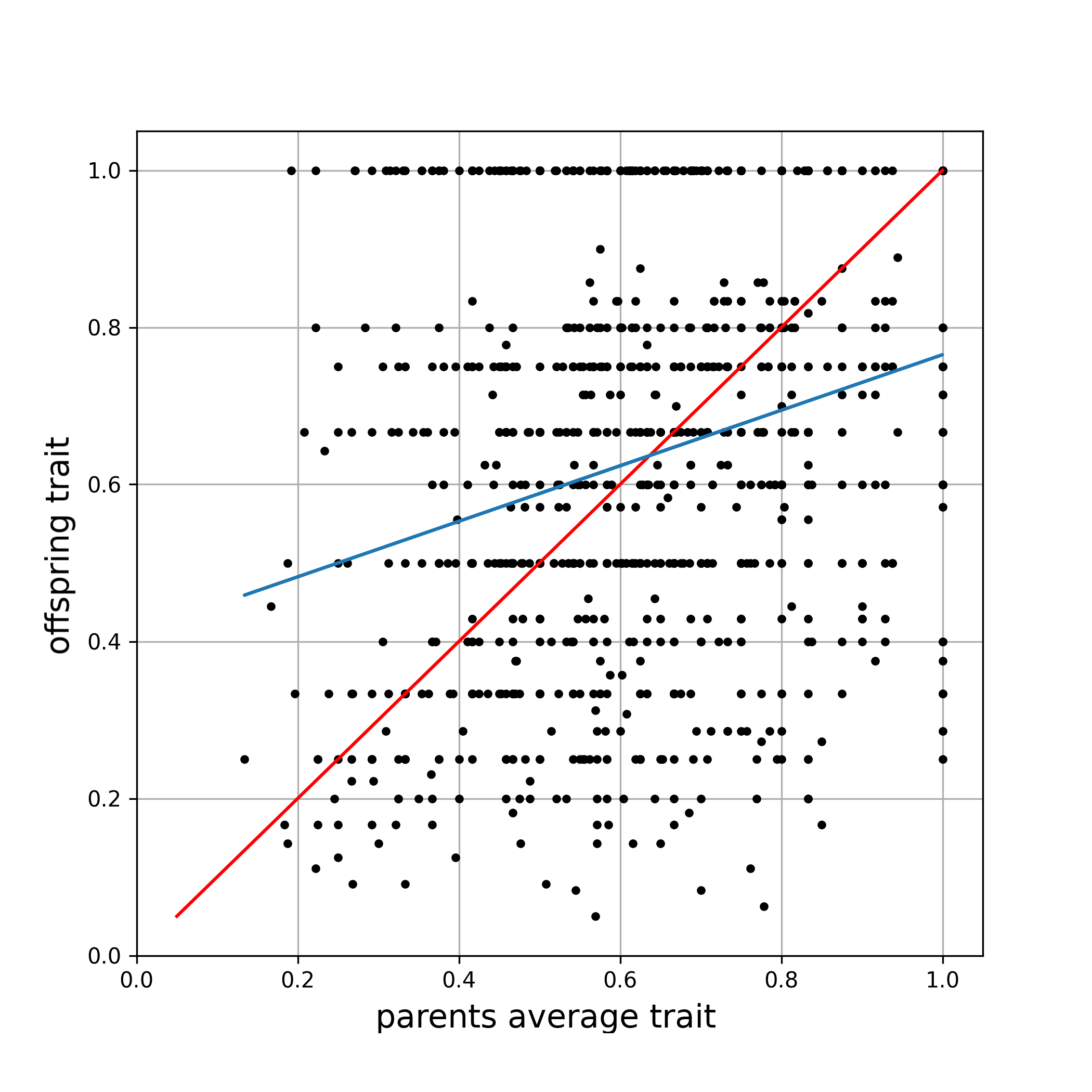}
        \caption{Proportion: L-system}
    \end{subfigure}
    \begin{subfigure}[b]{\scattersize\linewidth}
        \includegraphics[width=\linewidth]{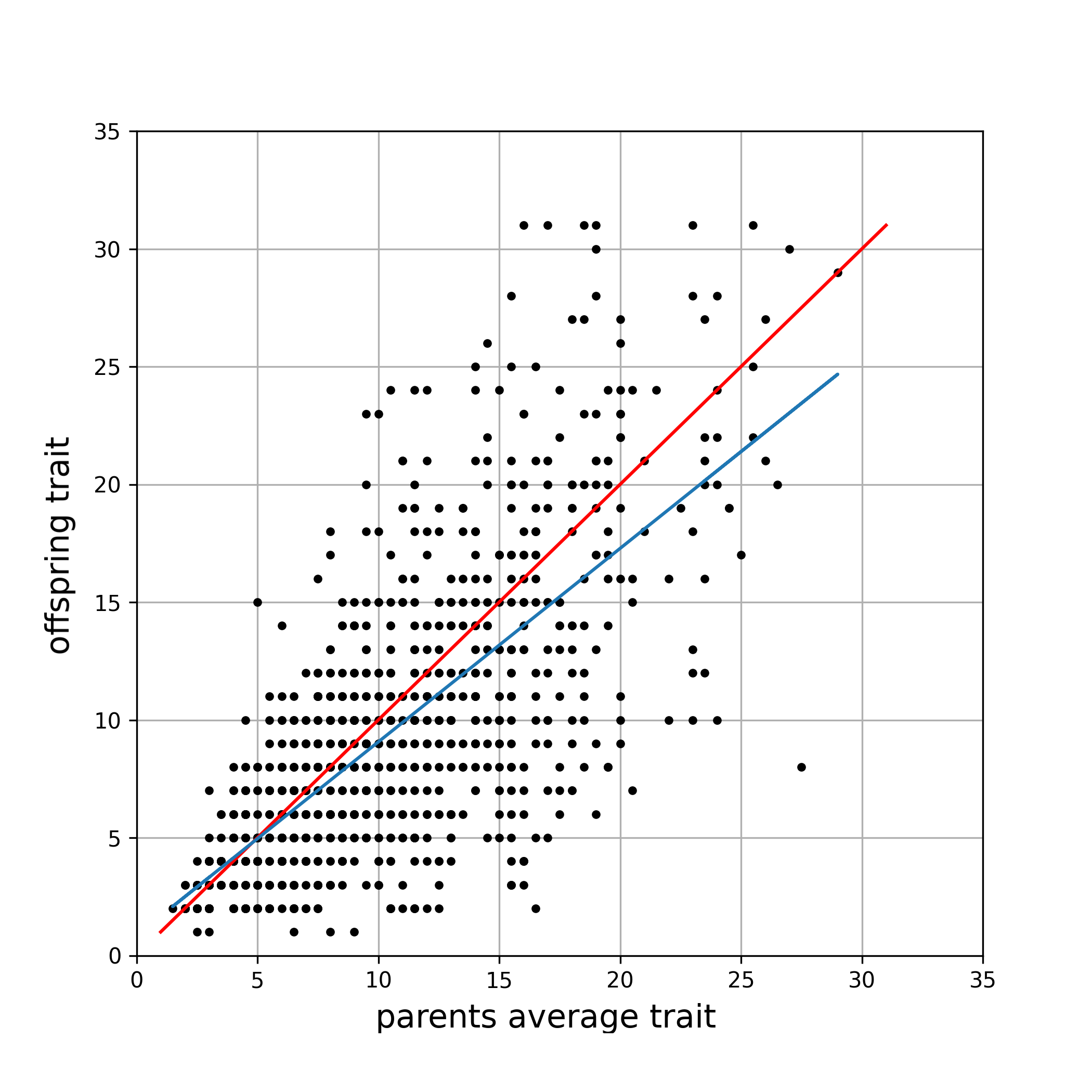}
        \caption{Size: L-system}
    \end{subfigure}
    \caption{Scatterplot for the two encodings for the parents in the first generation and their offspring. The x axis shows the average trait value of the parents, the y axis shows the value of the offspring. The blue line represents the linear regression of these values. A steeper line indicates a higher level of heritability, to a theoretical maximum of 1 ($45\degree$ slope). The red line is a $45\degree$ reference line.}
    \label{fig:heritability_scatters}
\end{figure*}

Our first analysis aims at measuring the heritability of various traits in the two encoding schemes at the start of the evolution experiment.
To do so, for each encoding/trait pair, we measure heritability using data from all evolutionary runs, but only on the very first generation.
Heritability is measured comparing the trait value of an offspring against the average of the parents, therefore we need the offspring from the second generation as well.
A linear regression is applied to the trait values of parents against offspring, and the slope of the resulting linear model is our estimate of heritability.
The estimated values from our measurements are reported in Table~\ref{tab:heritability_all_values}.
The scatter plots used for estimating heritability can be seen in Figures~\ref{fig:speed:scatter_tree}, \ref{fig:speed:scatter_lsystem}, \ref{fig:limbs:scatter_tree}, \ref{fig:limbs:scatter_lsystem} and Figure~\ref{fig:heritability_scatters}. As we can see, the Tree-based encoding consistently shows higher values of heritability for all traits.

\subsection{Relationship between heritability and initial evolutionary response}
We aim at analyzing the relationship between heritability and evolutionary response for the two encoding schemes. To achieve this, we analyze one behavioral and one morphological trait using a common scheme, as used in Figure~\ref{fig:speed} and Figure~\ref{fig:limbs}. Namely, in Panels (a) and (b) of the figures we will compare the heritability for direct vs indirect encoding calculated only in the first generation. In Panels (c) and (e) we show the dynamics of the trait being considered. Panel (c) shows the value of the trait over generations while Panel (e) shows its rate of change (or derivative) over generations. To further understand how heritability can explain the trait dynamics over generations, we use Panels (d) to highlight the evolution over generations of the heritability metric (calculated across two consecutive generations) and Panel (f) to highlight the evolution over a generation of the phenotypic diversity of the trait being considered within the population.

\begin{figure}[b]
    \centering
    \includegraphics[width=.85\linewidth]{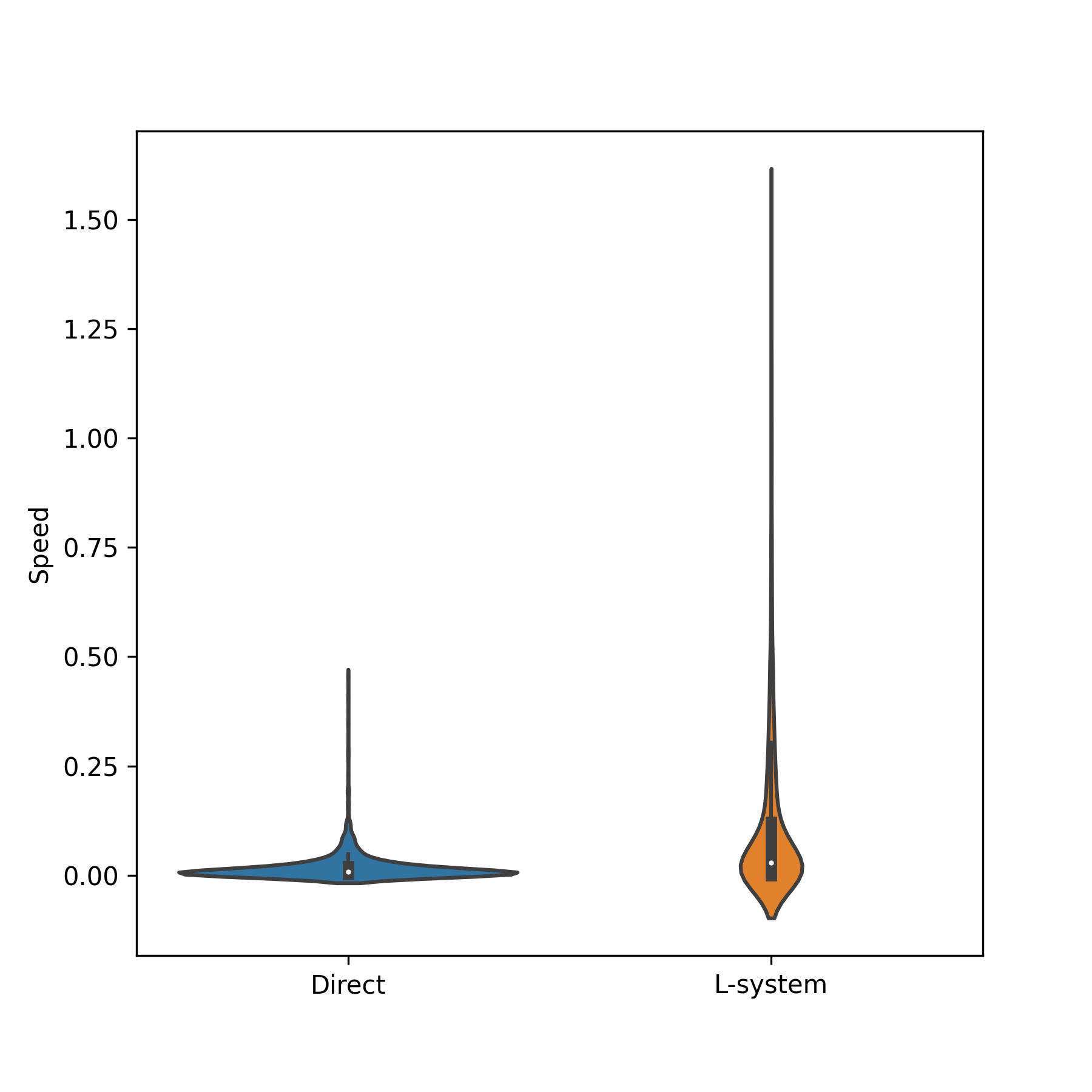}
    \caption{\small Distribution of fitness in the random initial population}
    \label{fig:fitness_gen0}
\end{figure}

We first analyze the most important trait, since this is the one that is under selection: speed.
In Figure~\ref{fig:speed:median} and Figure~\ref{fig:speed:derivate}, we observe that in the first $10$ generations of the evolutionary process the Tree-based representation has a higher rate of change in fitness compared to the L-System.
This finding is further confirmed by looking at the fitness distribution at Generation $0$ for all runs and for the two encodings, shown in Figure~\ref{fig:fitness_gen0}. Here we see that the L-System even starts with an advantage, represented by the much higher fitness diversity in the initial population and the corresponding presence of higher fitness individuals. Despite this, the L-System experiment evolves initially at a slower rate than the Tree-based experiment (as in Figure~\ref{fig:speed:derivate}, which means that those high-fitness individuals present in the populations are not able to pass their phenotype to their offspring to the same extent as it happens in the Tree-based representation.
We argue that the concept of heritability can shed light to understand better what we observed above. Indeed, the Tree-based encoding has a higher heritability value than the L-System encoding (see Figure~\ref{fig:speed:scatter_tree} and Figure~\ref{fig:speed:scatter_lsystem}).
Heritability can inform us on how much of the phenotypic trait variation will be passed on from parents to offspring, therefore high heritability at the beginning of the evolutionary process can predict a higher rate of change in the trait under selection, as is happening in our system.
Thus, higher heritability at the beginning of the evolutionary process directly facilitates the effect of initial selection, because good parents have a higher probability of creating good offspring.

Importantly, the relation between initial heritability and the initial rate of change of a trait is not true only for traits that are specifically under selection. To support this claim, we perform the same analysis for all traits, and we report here only one example.
Figure~\ref{fig:limbs:derivate} shows the rate of change of a morphological trait that is not under selection (number of limbs).
Also here, we observe the same overall pattern: the Tree-based representation has a higher initial rate of change in this trait, consistently with having higher heritability (Table~\ref{tab:heritability_all_values}). 

\subsection{Heritability during the later phase of evolution}
The analysis of heritability can not only help to describe the behavior of evolutionary systems in their initial phases, but it can also show interesting patterns during later advanced phases of their process.
From the theoretical definition, we expect the estimated value of heritability to be constant over the course of evolution, under the condition that the selection process does not affect the genetic variation in the population.
Surprisingly, by computing the estimated value of heritability for each of the generations, shown in the top row of Figure~\ref{fig:heritability_diversity}, we observe changes of the estimated heritability in our experiments.
The change in heritability value across generations is most evident for traits that are not under selection.
When looking at
Figure~\ref{fig:heritability_diversity}, we observe in the first generations a tendency for heritability to decrease for the Tree-based representation and to increase for the L-System representation.
In the later stages of evolution heritability stabilizes for the tree-based representation and becomes highly unstable for the L-system genotype.
\begin{figure}
    \centering
    \includegraphics[width=.72\linewidth]{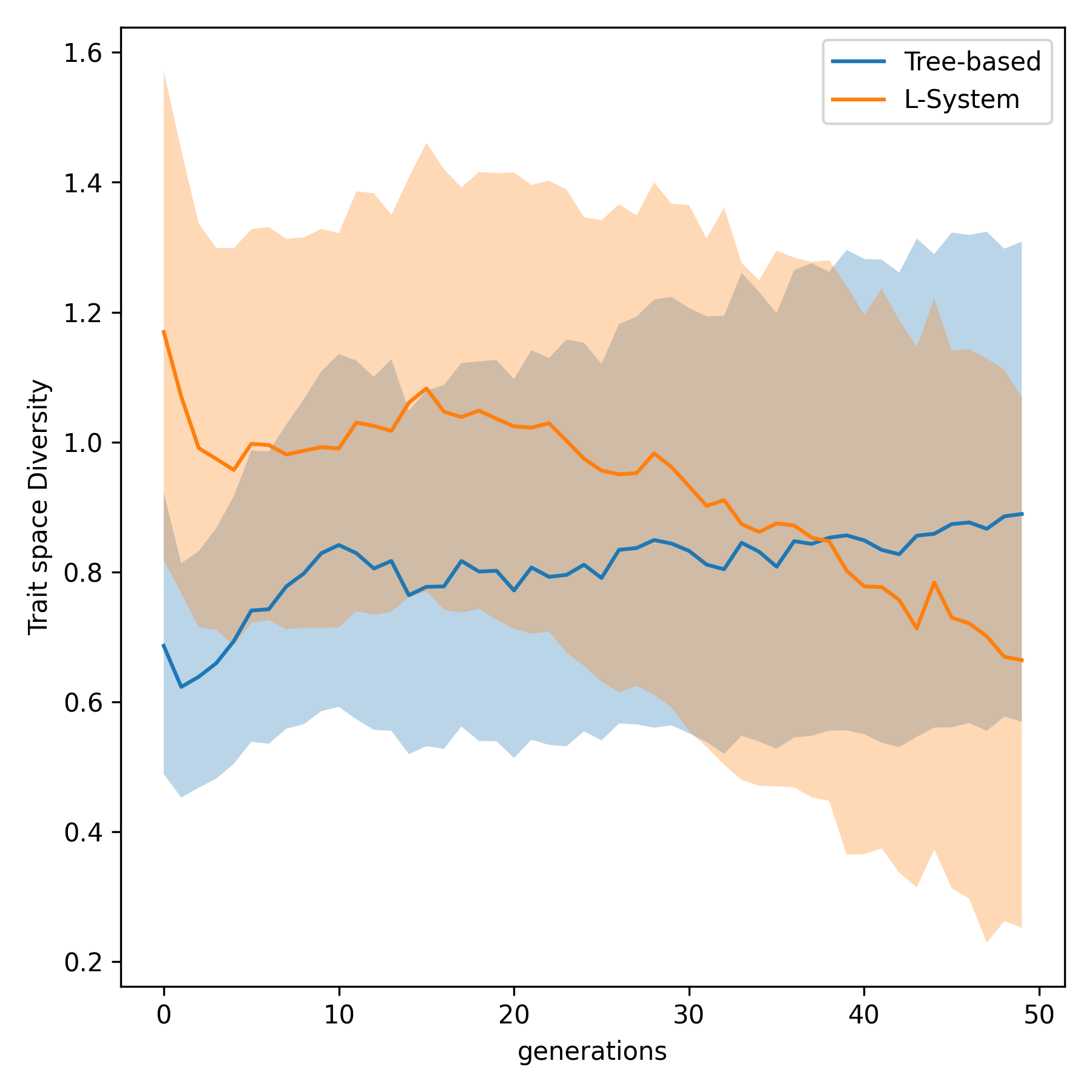}
    \caption{\small Diversity per generation. Diversity is the average distance of one individual against all other individuals in the same population. Distance is calculated in the trait space.}
    \label{fig:overall_diversity}
\end{figure}
The change of heritability over generations can be explained if we also analyze how the phenotypic diversity of the population over the generations varies.
The diversity in the L-system population converges to zero quite quickly in all traits (bottom row of Figure~\ref{fig:heritability_diversity}), except the one we select for: speed (Figure~\ref{fig:speed:diversity}).
In Figure~\ref{fig:overall_diversity} we can confirm and overall loss of phenotypic diversity for L-system experiments.
An overall decrease in phenotypic diversity in artificial evolutionary systems is often observed when evolution is converging to a solution and it is caused by a corresponding loss of genotypic diversity.
A change in genotypic diversity can also explain the change in heritability we measured.
A similar pattern can be observed in the tree-based representation, but the changes in heritability and diversity are visible on a smaller scale, for fewer generations and smaller changes in values.

Interestingly, we observed an unexpected overall increase in heritability for the L-system, as the value for heritability started relatively low and increased over generations.
We hypothesize that selection is responsible for this effect.
The mating selection is probably slowly excluding all individuals that present highly unpredictable gene sequences.
These highly unpredictable gene sequences can cause very poor offspring to be generated from very fit parents and vice-versa.
Slowly the selection process would pick up these highly unpredictable gene sequences in their low-fitness state and select them out of the next generation.
This effect throughout many generations would explain a decrease in diversity and an increase in narrow-sense heritability, as only the predictable gene sequences consistently survive across multiple generations and predictable gene sequences have cause high heritability in the population by definition.

\begin{figure}
    \centering
    \includegraphics[width=.75\linewidth]{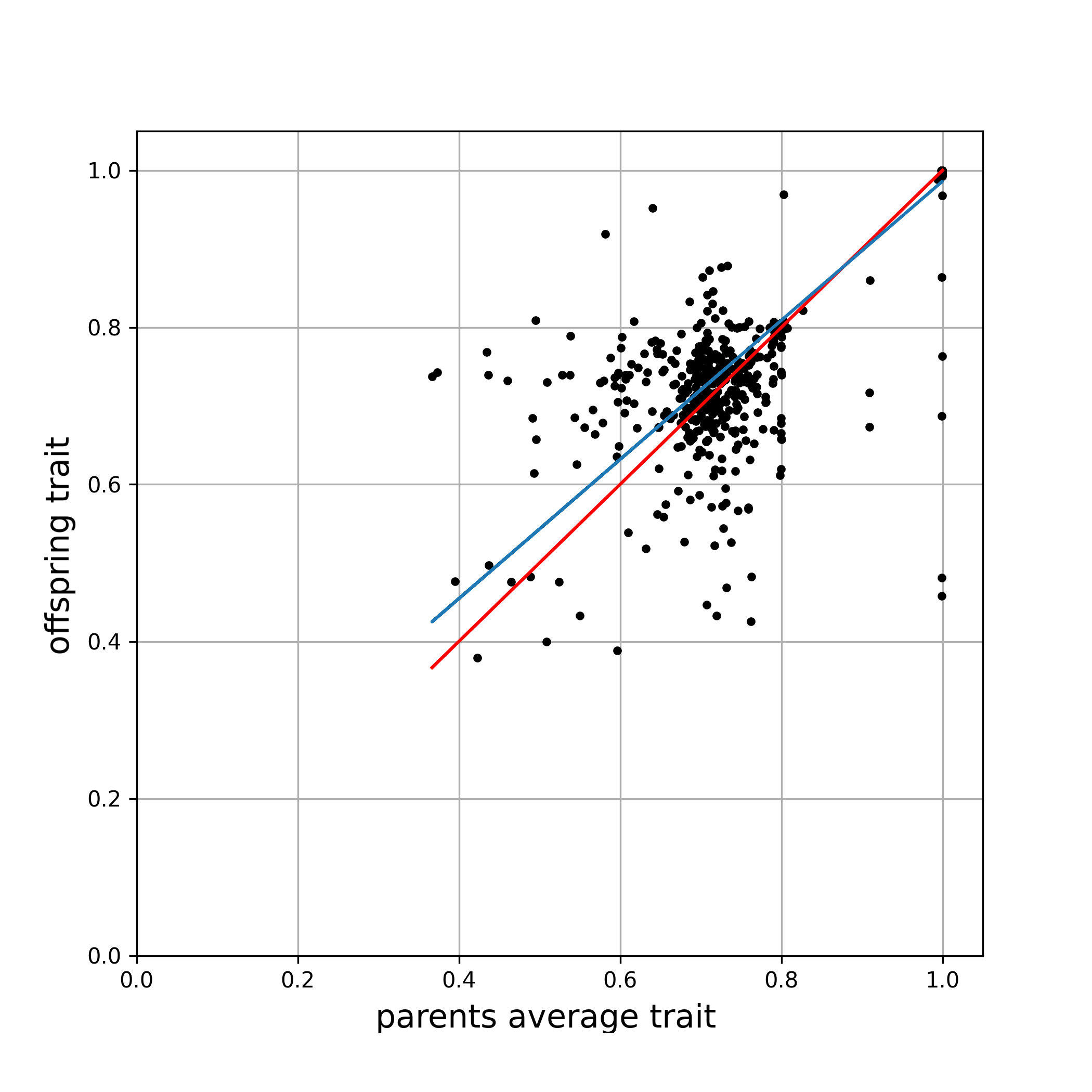}
    \caption{\small Scatter plot and heritability for balance of the last generation for the L-system configuration. The x axis is the average balance of the parents and the y axis is the value of balance for the offspring.  The blue line represents the linear regression of these values. A steeper line indicates a higher level of heritability, to a theoretical maximum of 1 ($45\degree$ slope). The red line is a $45\degree$ reference line. The linear regression results to almost a $45\degree$ line because there are many overlapping points in the plot that indistinguishable but influence the regression results.}
    \label{fig:balance:heritability_lastgen_lsystem}
\end{figure}

Another pattern that we can observe is that heritability becomes highly unstable in later stages of evolution for L-system experiments; this is especially obvious in
Figure~\ref{fig:balance:heritability_generation}.
The explanation for this effect can be found by looking at the diversity (Figure~\ref{fig:balance:diversity}).
A decrease in diversity for a trait is an indicator of a decrease in overall genotypic diversity, which implies that the evolutionary process is not exploring the search space any more and all solutions are very similar.
But to compute an accurate estimation of heritability we need high diversity of the population, otherwise we are computing the linear regression of a concentrated cloud of points, as shown in Figure~\ref{fig:balance:heritability_lastgen_lsystem}.
By contrast, we deduce that the tree-based experiments are exploring the evolutionary space and having difficulties in exploiting the solution space.
With this knowledge we can estimate that the tree-based representation needs a lower mutation rate and a stronger selection pressure to be able to exploit the solution space.
This was to be expected because we choose very high mutation rates and a relaxed selection mechanism.
On the contrary L-system seems to drop diversity as early as generation 20, which is pretty early.
This is especially noticeable for trait which we don't select for (Figures~\ref{fig:balance:diversity},~\ref{fig:proportion:diversity},~\ref{fig:size:diversity}).
Our L-system encoding seem to have difficulties to explore even from the early phases of evolution, despite using evolutionary parameters that encourage exploration.
This suggest that an evolutionary process using our L-system encoding has a hard time escaping a few local optima found at the beginning, and probably needs some additional elements to encourage exploration.


\newcommand{\divsize}{0.325}
\begin{figure*}
    \centering
    \begin{subfigure}[b]{\divsize\linewidth}
        \centering
        \includegraphics[width=\linewidth]{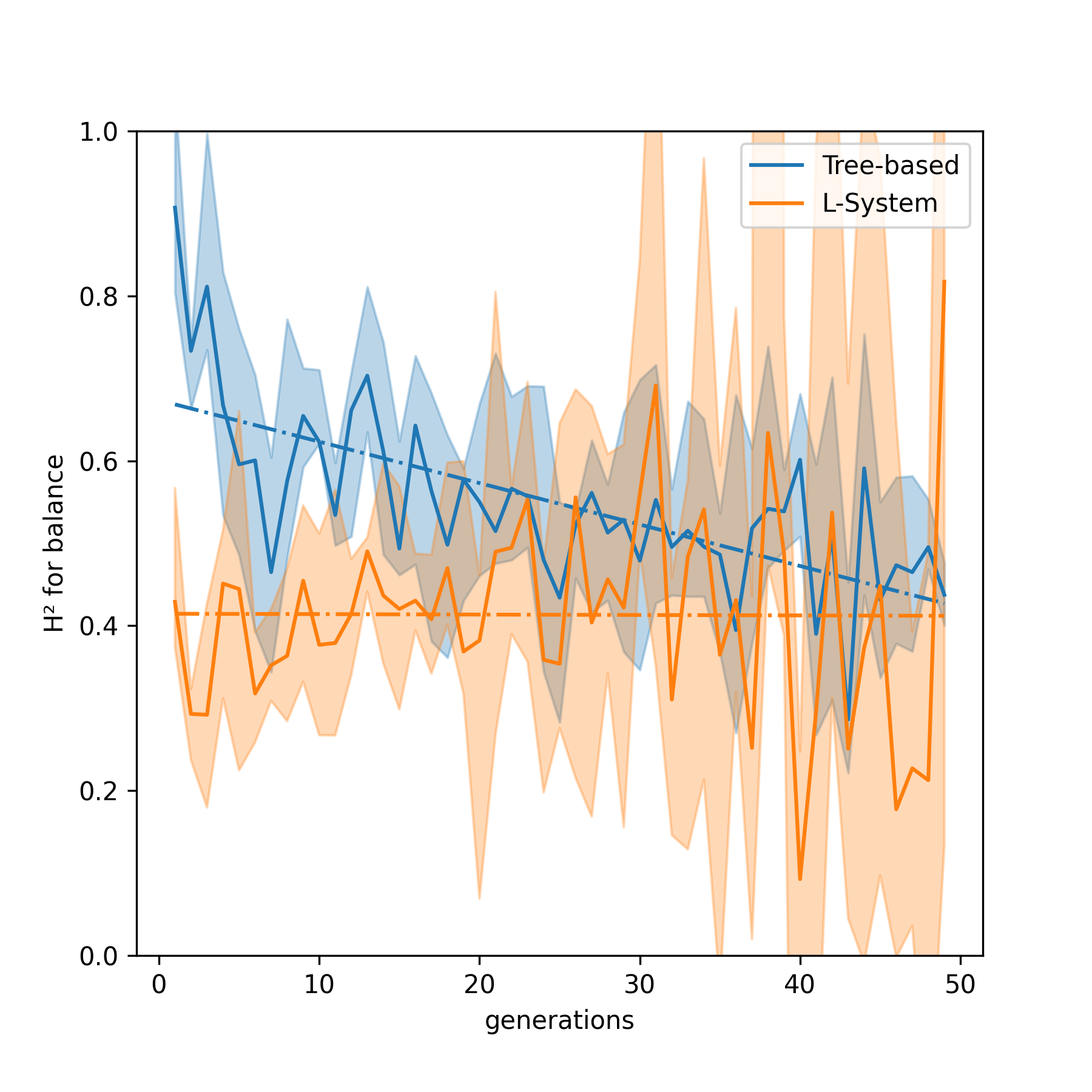}
        \caption{\footnotesize Balance Heritability}
        \label{fig:balance:heritability_generation}
        \vspace{.3cm}
    \end{subfigure}
    \hfill
    \begin{subfigure}[b]{\divsize\linewidth}
        \centering
        \includegraphics[width=\linewidth]{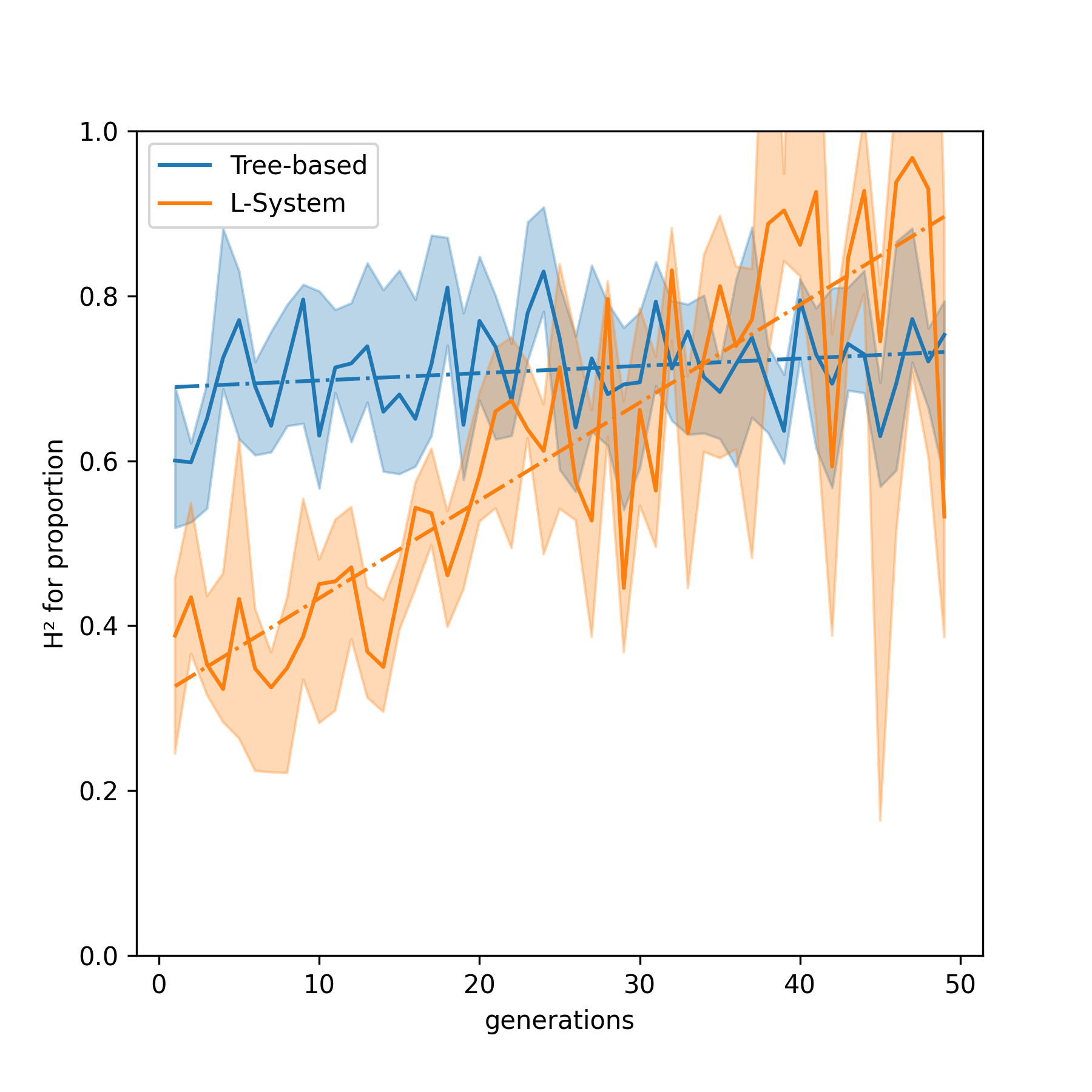}
        \caption{\footnotesize Proportion Heritability}
        \label{fig:proportion:heritability_generation}
        \vspace{.3cm}
    \end{subfigure}
    \hfill
    \begin{subfigure}[b]{\divsize\linewidth}
        \centering
        \includegraphics[width=\linewidth]{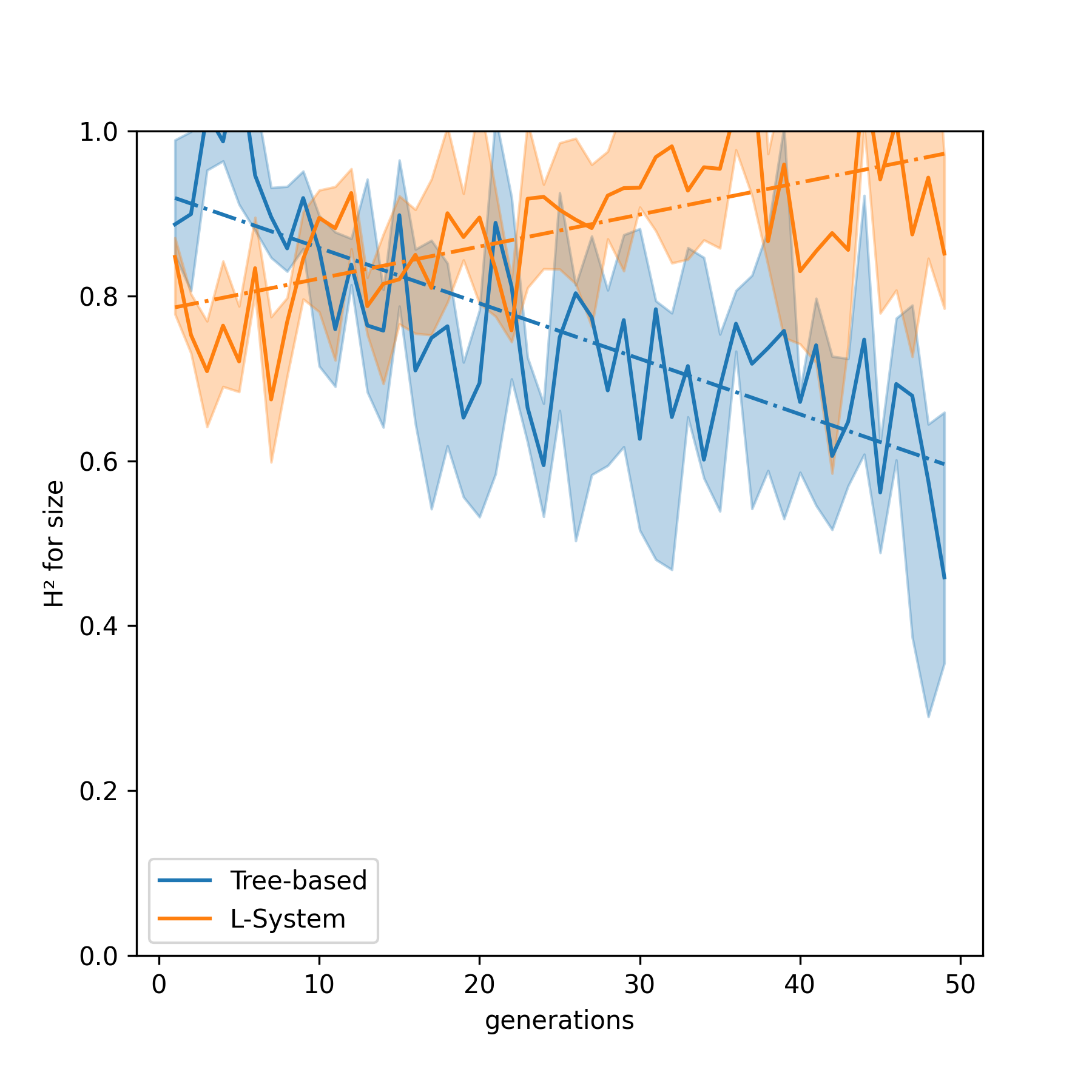}
        \caption{\footnotesize Size Heritability}
        \label{fig:size:heritability_generation}
        \vspace{.3cm}
    \end{subfigure}
    \begin{subfigure}[b]{\divsize\linewidth}
        \centering
        \includegraphics[width=.95\linewidth]{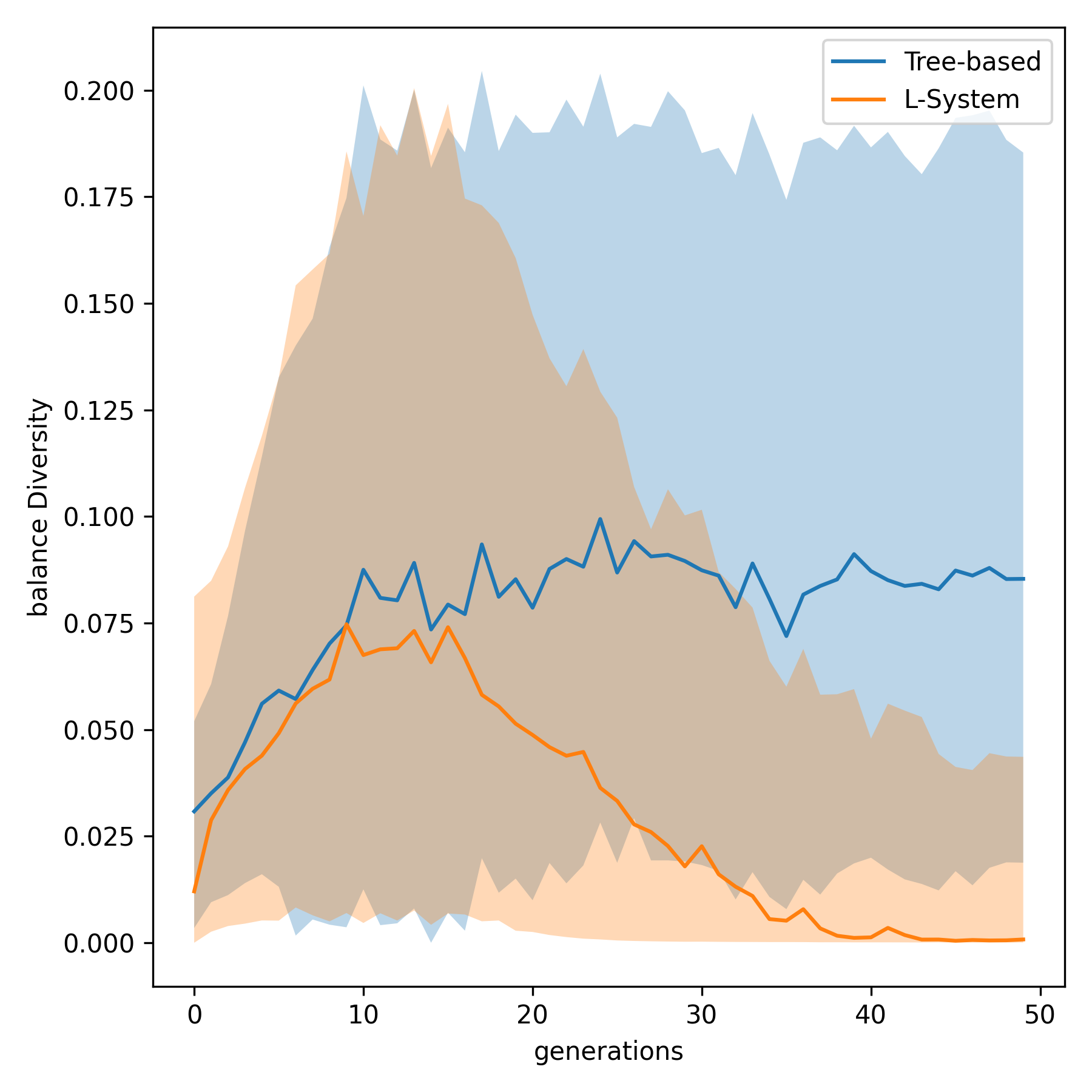}
        \caption{\footnotesize Median diversity of balance per generation}
        \label{fig:balance:diversity}
    \end{subfigure}
    \hfill
    \begin{subfigure}[b]{\divsize\linewidth}
        \centering
        \includegraphics[width=.95\linewidth]{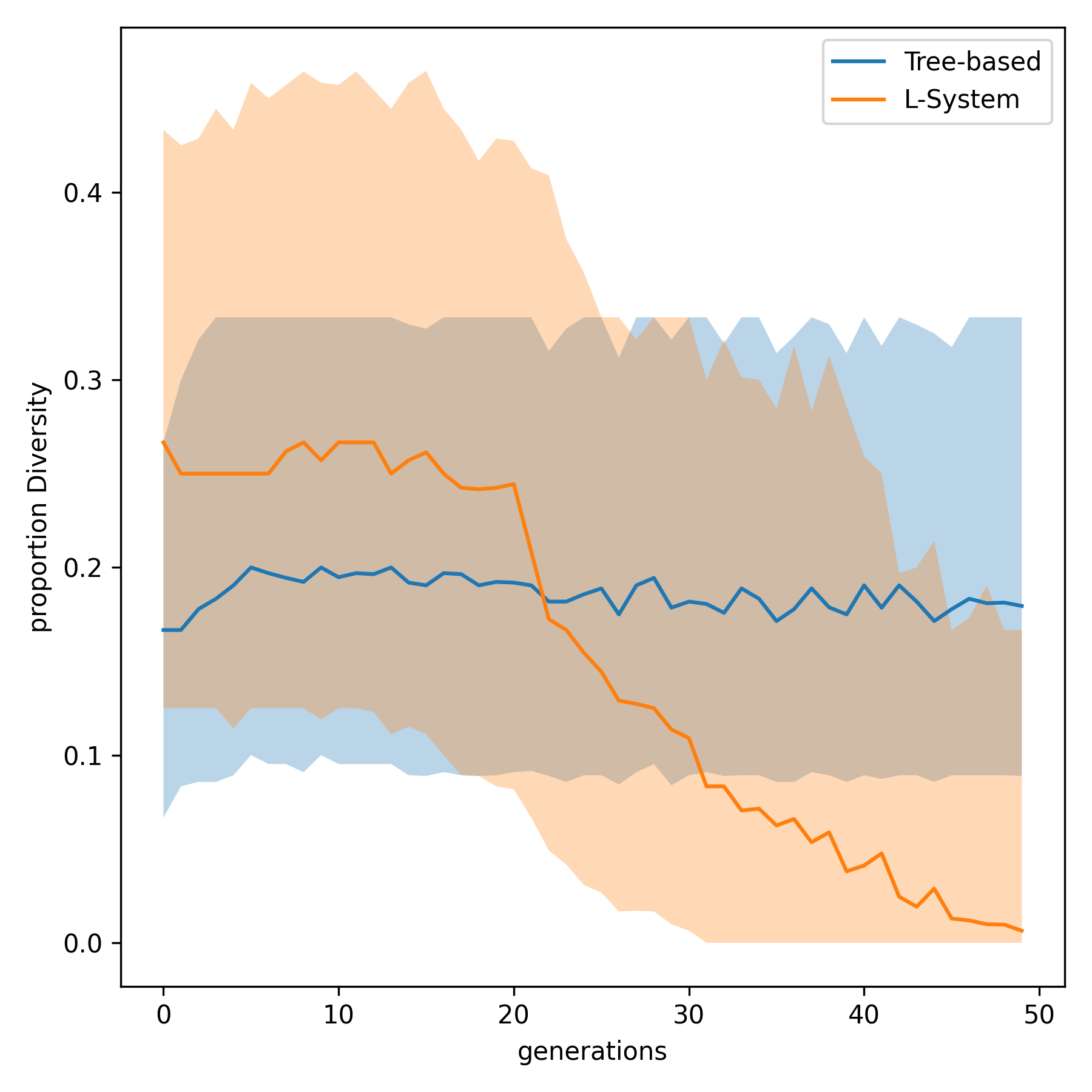}
        \caption{\footnotesize Median diversity of proportion per generation}
        \label{fig:proportion:diversity}
    \end{subfigure}
    \hfill
    \begin{subfigure}[b]{\divsize\linewidth}
        \centering
        \includegraphics[width=.95\linewidth]{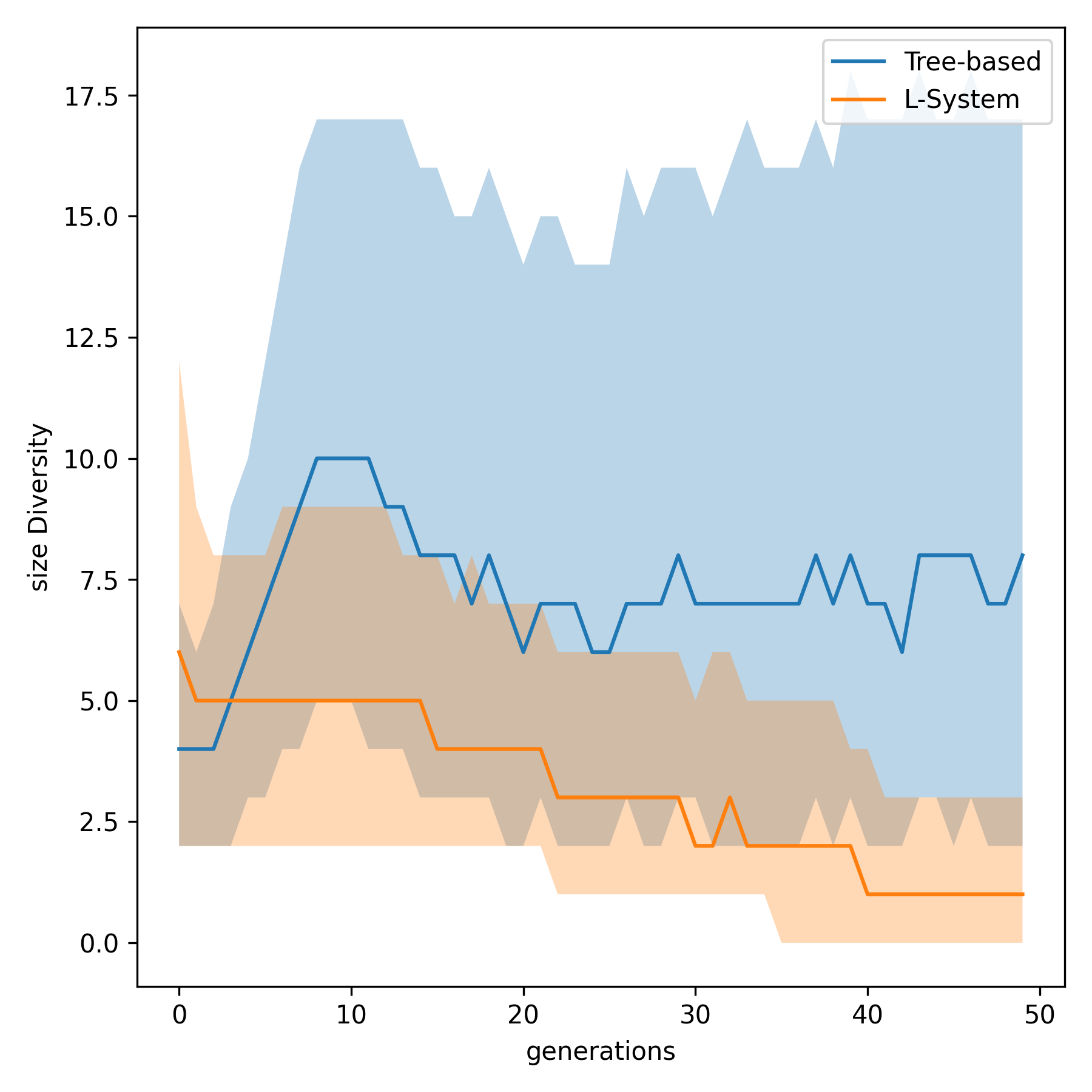}
        \caption{\footnotesize Median diversity of size per generation}
        \label{fig:size:diversity}
    \end{subfigure}
    \vspace{.3cm}
    \caption{plot of Heritability and Diversity over generations for three different phenotypic traits: Balance, Proportion and Size}
    \label{fig:heritability_diversity}
\end{figure*}


\section{Concluding remarks}

In this paper we introduced the biological notion of heritability as a novel tool to study encodings in evolutionary robotics.
Heritability captures the correlation between a quantifiable phenotypical trait measured in the parents and the one measured in the offspring.
In our experiments we show that heritability can be a useful tool in evolutionary robotics to support the genotype design process.
We used this novel tool to tackle the bootstrapping problem, because it reveals how exploratory a system is during the initial phases of evolution.
We observed how towards the course of evolution, changes in heritability could correlate to changes in diversity; i.e. in our tree-based system diversity and heritability seem to stabilize, while, in the L-system experiments, diversity drops and the estimated heritability increases at first following by high instability.
We related the different rates of heritability and diversity to different behaviours of exploration and exploitation and how these concepts seem to be intertwined, i.e. we observed a correlation between exploratory behaviours and high narrow-sense heritability, and inverse correlation between exploitative behaviours and low heritability, caused by a greater epistasis effect in the encoding.
Importantly, this analysis can be performed only within the first few generations (in our case, $50$), during the transitory phase of the evolutionary process. 

Heritability proved to be a helpful tool to evaluate the shape and smoothness of the search-space, considering the landscape of both fitness and other phenotypic traits.
Tree-based experiments converge to solutions where robots still retain significant morphological diversity, meaning the local optima found by evolution in the search-space is a smooth wide hill.
This was expected by a genetic encoding that is mostly made of additive genes.
On the contrary, L-system experiments converged to a single morphological solution with little-to-none diversity, and by observing the change in heritability and diversity we can determine that solutions are unlikely to explore other peaks, probably because they are either too distant, too narrow or not good enough compare to the solution found.
This is an indication that the L-system is a genetic encoding that contains lots of epistatic effects, meaning many genes need to align to be able to create a positive effect on the phenotype.

In possess of this knowledge, we will transition to a tree-based direct encoding in future work where the focus will be on evolution of morphological traits interacting with other elements, because differences in phenotypical traits will be more tangible.
In other setups, were the interest is in studying complicated gene interaction, including epistatic effects, L-system is a good candidate.

This paper also highlights how the notion of heritability draws attention to major discrepancies between biological and artificial evolutionary systems.
Crucially, in biology computing a value of heritability requires measuring qualities of the traits at the phenotype level.
However, phenotypic traits in biology are a result of several processes and factors: recombination, mutation, embryonic development, early life development, behavioural and morphological lifetime adaptation.
Many of these processes can be influenced by complex environments.
In addition, biological systems have and retain high genotypic variation.

In contrast, artificial systems are extremely simplified.
In particular, in our evolutionary system, individuals develop before they can have any interaction with the environment. Additionally, when deployed in the environment, individuals have no morphological or behavioural adaptation systems at their disposal, no \emph{development}.
In addition, our environment offers minimal interactions with the environment.
Artificial systems are also characterized by a generally lower genotypic diversity and a tendency to decrease even more over the course of artificial evolution, when the population converges to a useful solution.


For the future of evolutionary robotics, adding the above-mentioned biological elements would be very interesting, and some efforts have already been done in this direction.
In \cite{castillo_evolving-controllers_2020, gupta_embodied_2021} we find efforts to introduce learning systems that enable individuals to adapt to their environment during their lifetime.
Some work can also be found studying how to design a system where the genotype-phenotype mapping can be influenced by the environment \cite{miras_environmental_2020}.
However, these additions to the evolutionary process are non-trivial:
research on these more complicated and realistic systems require a substantial increase in computational cost required from the evolutionary process.
This results in slower iterations and difficulties in the design and parameter tuning processes, especially if one wanted to study  these processes in combination.

The development of artificial life through artificial evolution is still in his infancy, and lot of work in the above directions and beyond could be done.
Still, both for the current system complexity as well as for the one of future systems, in our view measuring heritability will be a useful tool that greatly increases our understanding of the relationship between phenotypes and genotypes.


\printbibliography

@book{lynch1998genetics,
  title={Genetics and analysis of quantitative traits},
  author={Lynch, Michael and Walsh, Bruce and others},
  volume={1},
  year={1998},
  publisher={Sinauer Sunderland, MA}
}

@article{dochtermann2019heritability,
  title={The heritability of behavior: a meta-analysis},
  author={Dochtermann, Ned A and Schwab, Tori and Anderson Berdal, Monica and Dalos, Jeremy and Royaut{\'e}, Rapha{\"e}l},
  journal={Journal of Heredity},
  volume={110},
  number={4},
  pages={403--410},
  year={2019},
  publisher={Oxford University Press US}
}

@article{mousseau1987natural,
  title={Natural selection and the heritability of fitness components},
  author={Mousseau, Timothy A and Roff, Derek A},
  journal={Heredity},
  volume={59},
  number={2},
  pages={181--197},
  year={1987},
  publisher={Nature Publishing Group}
}

@inproceedings{carlo_influences_2020,
	location = {Canberra, {ACT}, Australia},
	title = {Influences of Artificial Speciation on Morphological Robot Evolution},
% 	isbn = {978-1-72812-547-3},
% 	url = {https://ieeexplore.ieee.org/document/9308433/},
% 	doi = {10.1109/SSCI47803.2020.9308433},
% 	eventtitle = {2020 {IEEE} Symposium Series on Computational Intelligence ({SSCI})},
	pages = {2272--2279},
	booktitle = {2020 {IEEE} Symposium Series on Computational Intelligence ({SSCI})},
	publisher = {{IEEE}},
	author = {Carlo, Matteo De and Zeeuwe, Daan and Ferrante, Eliseo and Meynen, Gerben and Ellers, Jacintha and Eiben, A.E.},
% 	urldate = {2021-07-19},
	date = {2020-12},
}

@incollection{castillo_evolving-controllers_2020,
	location = {Cham},
	title = {Evolving-Controllers Versus Learning-Controllers for Morphologically Evolvable Robots},
	volume = {12104},
	pages = {86--99},
	booktitle = {Applications of Evolutionary Computation},
	publisher = {Springer International Publishing},
	author = {Miras, Karine and De Carlo, Matteo and Akhatou, Sayfeddine and Eiben, A. E.},
	editor = {Castillo, Pedro A. and Jiménez Laredo, Juan Luis and Fernández de Vega, Francisco},
	year = {2020},
}

@article{lindenmayer_mathematical_1968,
	title = {Mathematical models for cellular interactions in development {II}. Simple and branching filaments with two-sided inputs},
	volume = {18},
	pages = {300--315},
	number = {3},
	journaltitle = {Journal of Theoretical Biology},
	author = {Lindenmayer, Aristid},
	year = {1968},
}

@misc{gupta_embodied_2021,
	title = {Embodied Intelligence via Learning and Evolution},
	howpublished = {{arXiv}:2102.02202 [cs]},
	author = {Gupta, Agrim and Savarese, Silvio and Ganguli, Surya and Fei-Fei, Li},
}

@inproceedings{miras_impact_2019,
	title = {The impact of environmental history on evolved robot properties},
	pages = {396--403},
	booktitle = {{ALIFE} 2019},
	publisher = {{MIT} Press},
	author = {Miras, Karine and Eiben, A.E.},
	year = {2019},
}

@inproceedings{miras_effects_2019,
	title = {Effects of environmental conditions on evolved robot morphologies and behavior},
	pages = {125--132},
	booktitle = {Proceedings of the 2019 Genetic and Evolutionary Computation Conference ({GECCO} 2019)},
	publisher = {{ACM} Press},
	author = {Miras, Karine and Eiben, A. E.},
	year = {2019},
}

@article{auerbach_environmental_2014,
	title = {Environmental Influence on the Evolution of Morphological Complexity in Machines},
	volume = {10},
	pages = {1--17},
	number = {1},
	journaltitle = {{PLoS} Computational Biology},
	author = {Auerbach, Joshua E. and Bongard, Josh C.},
	editor = {Sporns, Olaf},
	year = {2014},
}

@article{Stanley2007,
	title = {Compositional pattern producing networks: A novel abstraction of development},
	volume = {8},
	pages = {131--162},
	number = {2},
	journaltitle = {Genetic Programming and Evolvable Machines},
	author = {Stanley, Kenneth O.},
	year = {2007},
}

@inproceedings{lan_directed_2018,
	title = {Directed Locomotion for Modular Robots with Evolvable Morphologies},
	pages = {476--487},
	booktitle = {Parallel Problem Solving from Nature -- {PPSN} {XV}},
	publisher = {Springer International Publishing},
	author = {Lan, Gongjin and Jelisavcic, Milan and Roijers, Diederik M and Haasdijk, Evert and Eiben, A.E.},
	year = {2018},
}

@article{miras_environmental_2020,
	title = {Environmental influences on evolvable robots},
	volume = {15},
% 	issn = {1932-6203},
% 	url = {https://dx.plos.org/10.1371/journal.pone.0233848},
% 	doi = {10.1371/journal.pone.0233848},
	abstract = {The field of Evolutionary Robotics addresses the challenge of automatically designing robotic systems. Furthermore, the field can also support biological investigations related to evolution. In this paper, we evolve (simulated) modular robots under diverse environmental conditions and analyze the influences that these conditions have on the evolved morphologies, controllers, and behavior. To this end, we introduce a set of morphological, controller, and behavioral descriptors that together span a multi-dimensional trait space. Using these descriptors, we demonstrate how changes in environmental conditions induce different levels of differentiation in this trait space. Our main goal is to gain deeper insights into the effect of the environment on a robotic evolutionary process.},
	pages = {1--23},
	number = {5},
	journaltitle = {{PLOS} {ONE}},
	shortjournal = {{PLoS} {ONE}},
	author = {Miras, Karine and Ferrante, Eliseo and Eiben, A. E.},
	editor = {Horvath, Denis},
	date = {2020-05},
}

@article{griffiths_quantifying_2000,
	title = {Quantifying heritability},
	abstract = {If a trait is shown to have some heritability in a population, then it is possible to quantify the degree of heritability. In Figure 25-3, we saw that the variation between phenotypes in a population arises from two sources. First, there are average differences between the genotypes; second, each genotype exhibits phenotypic variance because of environmental variation. The total phenotypic variance of the population (S2p) can then be broken into two parts: the variance between genotypic means (S2g) and the remaining variance (S2e) The former is called the genetic variance, and the latter is called the environmental variance; however, as we shall see, these names are quite misleading. Moreover, the breakdown of the phenotypic variance into the sum of environmental and genetic variance leaves out the possibility of some covariance between genotype and environment. For example, suppose it were true (we do not know) that there are genes that influence musical ability. Parents with such genes might themselves be musicians, who would create a more musical environment for their children, who would then have both the genes and the environment promoting musical performance. The result would be an increase in the phenotypic variances of musical ability and an erroneous estimate of genetic and environmental variances. If the phenotype is the sum of a genetic and an environmental effect, P = G + E, then, as explained on page 768 of the Statistical Appendix, the variance of the phenotype is the sum of the genetic variance, the environmental variance, and twice the covariance between the genotypic and environmental effects.},
% 	pages = {Available at: http://www.ncbi.nlm.nih.gov/books/NB},
	journaltitle = {An Introduction to Genetic Analysis},
	author = {Griffiths, {AJF} and Miller, {JH} and Suzuki, {DT} et al},
	year = {2000},
}

@article{wray2008estimating,
	title = {Estimating trait heritability},
	volume = {1},
	pages = {29},
	number = {1},
	journaltitle = {Nature education},
	author = {Wray, Naomi and Visscher, Peter},
	year = {2008},
}

@inproceedings{de_carlo_comparing_2020,
	title = {Comparing Indirect Encodings by Evolutionary Attractor Analysis in the Trait Space of Modular Robots},
	pages = {73--74},
	booktitle = {Proceedings of the 2020 Genetic and Evolutionary Computation Conference ({GECCO} 2020)},
	author = {De Carlo, Matteo and Ferrante, Eliseo and Eiben, A.E.},
	year = {2020},
}

@article{stanley_hypercube-based_2009,
	title = {A hypercube-based encoding for evolving large-scale neural networks},
	volume = {15},
	pages = {185--212},
	number = {2},
	journaltitle = {Artificial Life},
	author = {Stanley, Kenneth O. and D'Ambrosio, David B. and Gauci, Jason},
	year = {2009},
}

@inproceedings{auerbach_robogen_2014,
	title = {{RoboGen}: Robot Generation through Artificial Evolution},
	pages = {136--137},
	booktitle = {Artificial Life 14: Proceedings of the Fourteenth International Conference on the Synthesis and Simulation of Living Systems},
	publisher = {{MIT} Press},
	author = {Auerbach, Joshua and Aydin, Deniz and Maesani, Andrea and Kornatowski, Przemyslaw and Cieslewski, Titus and Heitz, Grégoire and Fernando, Pradeep and Loshchilov, Ilya and Daler, Ludovic and Floreano, Dario},
	year = {2014},
}

@article{gauci_autonomous_2010,
	title = {Autonomous evolution of topographic regularities in artificial neural networks},
	volume = {22},
	pages = {1860--1898},
	number = {7},
	journaltitle = {Neural Computation},
	author = {Gauci, Jason and Stanley, Kenneth O.},
	year = {2010},
	pmid = {20235822},
}

@online{lan_learning_2020,
	title = {Learning Directed Locomotion in Modular Robots with Evolvable Morphologies},
	author = {Lan, Gongjin and Carlo, Matteo De and Diggelen, Fuda Van and Tomczak, Jakub M and Roijers, Diederik M and Eiben, A E},
	year = {2020},
	eprinttype = {arxiv},
	eprint = {2001.07804},
	note = {Publisher: {arXiv}},
}

@article{jelisavcic_lamarckian_2019,
	title = {Lamarckian Evolution of Simulated Modular Robots},
	volume = {6},
	pages = {1--15},
	issue = {February},
	journaltitle = {Frontiers in Robotics and {AI}},
	author = {Jelisavcic, Milan and Glette, Kyrre and Haasdijk, Evert and Eiben, A E},
	year = {2019},
}

@article{jelisavcic_real-world_2017,
	title = {Real-World Evolution of Robot Morphologies: A Proof of Concept},
	volume = {23},
	pages = {206--235},
	number = {2},
	journaltitle = {Artificial Life},
	author = {Jelisavcic, Milan and De Carlo, Matteo and Hupkes, Elte and Eustratiadis, Panagiotis and Orlowski, Jakub and Haasdijk, Evert and Auerbach, Joshua E. and Eiben, A. E.},
	year = {2017},
}

@article{hupkes_revolve:_2018,
	title = {Revolve: A Versatile Simulator for Online Robot Evolution},
	volume = {10784 {LNCS}},
	pages = {687--702},
	journaltitle = {Lecture Notes in Computer Science (including subseries Lecture Notes in Artificial Intelligence and Lecture Notes in Bioinformatics)},
	author = {Hupkes, Elte and Jelisavcic, Milan and Eiben, A. E.},
	year = {2018},
}

\end{document}